\documentclass[lettersize,journal]{IEEEtran}
\usepackage{amsmath,amsfonts}
\usepackage{algorithmic}
\usepackage{algorithm}
\usepackage{array}
\usepackage[caption=false,font=normalsize,labelfont=sf,textfont=sf]{subfig}
\usepackage{textcomp}
\usepackage{stfloats}
\usepackage{overpic}
\usepackage{epstopdf}
\usepackage{verbatim}
\usepackage{graphicx}
\usepackage{cite}
\usepackage[hyphens]{url} % 为了下标自动换行
\usepackage{color}
\usepackage{tikz}
\usetikzlibrary{shadows}
\usetikzlibrary{trees,positioning,shapes,shadows,arrows}
\usepackage[edges]{forest}
\usepackage[breaklinks,colorlinks,linkcolor=black,citecolor=black,urlcolor=black]{hyperref}
\usepackage[figuresright]{rotating}
\usepackage{booktabs}
\definecolor{hiddendraw}{RGB}{205, 44, 36}
\usepackage{makecell}

\usepackage{soul,color,xcolor}      % 用于设置字体颜色
\definecolor{myColor}{rgb}{0,0,0}        % 隐藏修订时用这行
\makeatletter
\newcommand*{\new}{\@ifnextchar\bgroup{\new@}{\color{myColor}}}
\newcommand*{\new@}[1]{{\textcolor{myColor}{#1}}}
\makeatother

\begin{document}

\title{Grid-Centric Traffic Scenario Perception for Autonomous Driving: A Comprehensive Review}

\author{Yining Shi$^{1,3}$, Kun Jiang$^{1*}$, Jiusi Li$^{1}$, Zelin Qian$^{1}$, Junze Wen$^{1}$,  Mengmeng Yang$^{1}$, Ke Wang$^{2}$ and Diange Yang$^{1*}$% <-this % stops a space
% \thanks{*This work was not supported by any organization}% <-this % stops a space
\thanks{$^{1}$Yining Shi, Kun Jiang, Jiusi Li, Zelin Qian, Junze Wen, Mengmeng Yang, and Diange Yang are with School of Vehicle and Mobility, Tsinghua University, Beijing, China. {\tt\small \{syn21,ljs23,qzl22, wjz22\}@mails.tsinghua.edu.cn, \{jiangkun,yangmm\_qh,ydg\}@mail.tsinghua.edu.cn}

$^{2}$Ke Wang is with Kargobot Inc. {\tt\small kewang@kargobot.ai}

Work partly done during Yining Shi's internship at DiDi Chuxing. Corresponding authors: Kun Jiang, Diange Yang.

}% <-this % stops a space
% \thanks{*This work was not supported by any organization}% <-this % stops a space
}

\markboth{preprint}%
{Shell \MakeLowercase{\textit{et al.}}: A Sample Article Using IEEEtran.cls for IEEE Journals}



\maketitle

% including classic occupancy grid maps (OGMs) to advanced occupancy networks. 

\begin{abstract}
Grid-centric perception is a crucial field for mobile robot perception and navigation. Nonetheless, grid-centric perception is less prevalent than object-centric perception as autonomous vehicles need to accurately perceive highly dynamic, large-scale traffic scenarios and the complexity and computational costs of grid-centric perception are high. In recent years, the rapid development of deep learning techniques and hardware provides fresh insights into the evolution of grid-centric perception. \new{The fundamental difference between grid-centric and object-centric pipeline lies in that grid-centric perception follows a geometry-first paradigm which is more robust to the open-world driving scenarios with endless long-tailed semantically-unknown obstacles. } Recent researches demonstrate the great advantages of grid-centric perception, such as comprehensive fine-grained environmental representation, greater robustness to occlusion and irregular shaped objects, better ground estimation, and safer planning policies. \new{There is also a growing trend that the capacity of occupancy networks are greatly expanded to 4D scene perception and prediction and latest techniques are highly related to new research topics such as 4D occcupancy forecasting, generative AI and world models in the field of autonomous driving.} Given the lack of current surveys for this rapidly expanding field, we present a hierarchically-structured review of grid-centric perception for autonomous vehicles. \new{We organize previous and current knowledge of occupancy grid techniques along the main vein from 2D BEV grids to 3D occupancy to 4D occupancy forecasting. We additionally summarize label-efficient occupancy learning and the role of grid-centric perception in driving systems.} Lastly, we present a summary of the current research trend and provide future outlooks.
\end{abstract}


\section{Introduction}
\IEEEPARstart{S}{afe} operation of autonomous vehicles requires an accurate and comprehensive representation of the surrounding environment. There are two mainstream perception pipelines for autonomous vehicles: object-centric and grid-centric pipelines. As a widely used pipline, object-centric pipeline, consists of 3D object detection, tracking, and trajectory prediction. Nevertheless, object-centric pipeline usually fails in open-world traffic scenarios where the appearance or pattern of objects is not well-defined for networks. These obstacles, also known as long-tail obstacles, include deformable obstacles, such as two-section trailers; special-shaped obstacles, such as overturned vehicles; obstacles of unknown categories, such as gravel on the road, garbage; partially obscured objects, etc. Thus, a more robust representation of these long-tail problems is urgently required.

As an emerging solution to cope with the endless open-world perception challenge, grid-centric pipeline is believed to be a promising solution, as it is able to provide occupancy and motion of any position in 3D surrounding space without knowing the specific category of an object. The significant advantages of grid-based representations over object-based representations are as follows: Insensitive to obstacle's geometric shape or semantic category, and stronger resistance to occlusion; A uniform space coordinate to align multi-modal sensor data; robust uncertainty estimation. However, the major drawbacks of grid-centric perception are the high computational complexity and label generation complexity, so label-efficient and computationally-efficient occupancy networks are more favorable to large-scale training and on-board deployment.

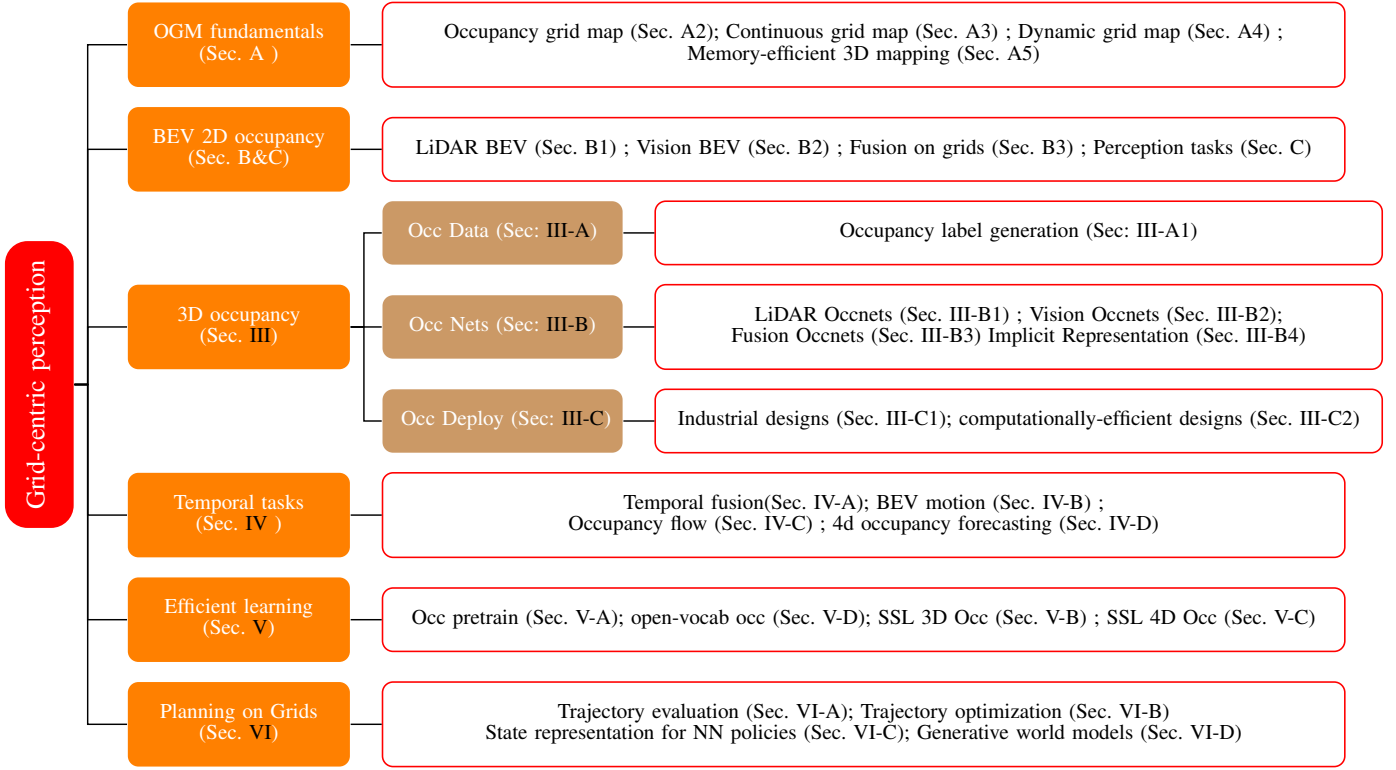
\begin{figure*}[htbp]
 \centering
\begin{forest}
  forked edges,
  for tree={
    grow=east,
    reversed=true,%increase counter-clockwise
    anchor=base west,
    parent anchor=east,
    child anchor=west,
    base=left,
    font=\small,
    rectangle,
    draw={hiddendraw, line width=0.6pt},
    rounded corners,align=left,
    minimum width=2.5em,
    edge={black, line width=0.55pt},
    l+=1.9mm,
    s sep=7pt,
    inner xsep=7pt,
    inner ysep=8pt,
    ver/.style={rotate=90, rectangle, draw=none, rounded corners=3mm, fill=red, text centered,  text=white, child anchor=north, parent anchor=south, anchor=center, font=\fontsize{10}{10}\selectfont,},
    level2/.style={rectangle, draw=none, fill=orange,  
    text centered, anchor=west, text=white, font=\fontsize{8}{8}\selectfont, text width = 7em},
    level3/.style={rectangle, draw=none, fill=brown,   fill opacity=0.8,text centered, anchor=west, text=white, font=\fontsize{8}{8}\selectfont, text width = 2.7cm, align=center},
    level3_2/.style={rectangle, draw=none, fill=gray,   fill opacity=0.8,text centered, anchor=west, text=white, font=\fontsize{8}{8}\selectfont, text width = 2.7cm, align=center},
    level3_1/.style={rectangle, draw=none, fill=brown,   fill opacity=0.8, text centered, anchor=west, text=white, font=\fontsize{8}{8}\selectfont, text width = 3.2cm, align=center},
    level4/.style={rectangle, draw=red, text centered, anchor=west, text=black, font=\fontsize{8}{8}\selectfont, align=center, text width = 2.9cm},
    level5/.style={rectangle, draw=red, text centered, anchor=west, text=black, font=\fontsize{8}{8}\selectfont, align=center, text width = 14.7em},
    level5_1/.style={rectangle, draw=red, text centered, anchor=west, text=black, font=\fontsize{8}{8}\selectfont, align=center, text width = 26.1em},
    level5_2/.style={rectangle, draw=red, text centered, anchor=west, text=black, font=\fontsize{8}{8}\selectfont, align=center, text width = 30.0em},
    level5_3/.style={rectangle, draw=red, text centered, anchor=west, text=black, font=\fontsize{8}{8}\selectfont, align=center, text width = 35.0em},
  },
  where level=1{text width=5em,font=\scriptsize,align=center}{},
  where level=2{text width=6em,font=\tiny,}{},
  where level=3{text width=6em,font=\tiny}{},%yshift=0.26pt
  where level=4{text width=5em,font=\tiny}{},%yshift=0.26pt
  where level=5{font=\tiny}{},%yshift=0.26pt
  [Grid-centric perception, ver
 %    [OGM fundamentals 
 % \\(Sec. \ref{appendix:ogm} ), level2  [Occupancy grid map (Sec. \ref{appendix:ogm_base}){;} Continuous grid map (Sec. \ref{appendix:com}) {;} Dynamic grid map (Sec. \ref{appendix:dogm}) , level5_3]]
 %    [BEV 2D grid \\
 % (Sec. \ref{sec:BEVRepre}), level2
 %        [LiDAR BEV (Sec. \ref{sec:lidar2bev}), level3
 %        ]
 %        [Vision BEV (Sec. \ref{sec:pv2bev}), level3
 %            [Geometry-based PV2BEV (Sec. \ref{sec:geo-pv2bev}){;} Network-based PV2BEV (Sec. \ref{sec:neural-pv2bev}) , level5_1]
 %        ]
 %        [Fusion (Sec. \ref{sec:deepfusion}), level3
 %            [Multi-sensor fusion (Sec. \ref{sec:multisensorfusion}){;} Multi-agent fusion (Sec. \ref{sec:multiagentfusion}) , level5_1]
 %        ]]
     [OGM fundamentals 
 \\(Sec. A ), level2  [Occupancy grid map (Sec. A2){;} Continuous grid map (Sec. A3) {;} Dynamic grid map (Sec. A4) {;} \\ Memory-efficient 3D mapping (Sec. A5), level5_3]]
       [BEV 2D occupancy \\
 (Sec. B\&C), level2 [LiDAR BEV (Sec. B1) {;} Vision BEV (Sec. B2) {;} Fusion on grids (Sec. B3) {;} Perception tasks (Sec. C) , level5_3]
 ]
         [3D occupancy \\
 (Sec. \ref{sec:3d_occu}), level2
        [Occ Data (Sec: \ref{sec:occ_data_piplie}), level3 [Occupancy label generation (Sec: \ref{sec:label_gen}), level5_1]]
        [Occ Nets (Sec: \ref{sec:occnets}), level3
            [LiDAR Occnets (Sec. \ref{sec:lidarssc}) {;} Vision Occnets (Sec. \ref{sec:visionssc}){;} \\Fusion Occnets (Sec. \ref{sec:fusionocc}) Implicit Representation (Sec. \ref{sec:implicitocc}) , level5_1]  
        ]
         [Occ Deploy (Sec: \ref{sec:deployocc}), level3 [Industrial designs (Sec. \ref{sec:industrialocc}){;} computationally-efficient designs (Sec. \ref{sec:compute-effi-occ}) , level5_1]]        
        ]
    [Temporal tasks
 \\(Sec. \ref{sec:temporal} ), level2  [Temporal fusion(Sec. \ref{sec:temporal_bev}){;} BEV motion (Sec. \ref{sec:short-term-motion}) {;} \\Occupancy flow (Sec. \ref{sec:occu_flow}) {;} 4d occupancy forecasting (Sec. \ref{sec:4docc}), level5_3]]
        [Efficient learning \\
 (Sec. \ref{sec:effilearn}), level2
            [Occ pretrain (Sec. \ref{sec:occ_as_pretrain}){;} open-vocab occ (Sec. \ref{sec:ov_occ}){;} SSL 3D Occ (Sec. \ref{sec:ssl_occ3d}) {;} SSL 4D Occ (Sec. \ref{sec:ssl_occ4d}), level5_3]
        ]
        % ]
            [Planning on Grids \\
 (Sec. \ref{sec:driveapplication}), level2
            [Trajectory evaluation (Sec. \ref{sec:grid_traj_eval}){;} Trajectory optimization  (Sec. \ref{sec:grid_traj_optim}) \\State representation for NN policies (Sec. \ref{sec:grid_state_repre}){;} Generative world models  (Sec. \ref{sec:simworld}) , level5_3]
        ]
]
\end{forest}
\new{
\caption{Hierarchically-structured taxonomy of grid-centric perception for autonomous driving.}
}
\label{fig:taxonomy}
\end{figure*}

\new{
In terms of the technological development path, there is a clear development trend from 2D modeling to 3D and to 4D modelling, aiming at a finer portrayal of spatio-temporally variable three-dimensional environments. Historically, occupancy grid mapping (OGM) has been widely acknowledged as an essential prerequisite for the safe navigation and collision avoidance of mobile robots and autonomous vehicles. With the development of deep neural networks, grid-centric methods are developing rapidly and now generate a more comprehensive understanding of semantics and motion than conventional OGM. Moreover, BEV-based networks are extended to full 3D occuapancy networks and further extended to 4D occupancy forecasting with temporal prediction modules.
}

Recent progress demonstrates that grid-centric perception is one of the emerging, promising and challenging research topics in autonomous vehicles. To this end, we intend to provide the first comprehensive review of grid-centric perception techniques.

\new{
\textbf{Comparison with existing surveys.} Existing surveys about automotive perception including 3D object detection\cite{3DODSurvey}, 3D object detection from images\cite{ImageODSurvey}, Neural Radiance Field (NeRF)\cite{NeRFSurvey},  vision-centric BEV perception\cite{BEVSurvey,DelvingBEV}, cover parts of the techniques in grid-centric perception. However, perception tasks, algorithms, and applications that center on grid representation are not thoroughly discussed in these evaluations. Latest preprints\cite{OccSurvey1, OccSurvey2} of reviews of occupancy networks only cover a limited field of grid-centric perception: 3D occupancy prediction from images and information fusion, respectively. 
}

\new{
In contrast, We present a comprehensive overview of grid-centric perception methods for autonomous vehicle applications, as well as an in-depth study and systematic comparison of grid-centric perception. In terms of feature representation, we cover  BEV 2D, 3D occupancy and 4D forecasting. In terms of data utility, we also investigate label-efficient learning for occupancy. In terms of driving system, We investigate how grid-centric perception benefits decision-making and planning. 
}

To sum up, the contributions of this survey paper is summarized as follows:

\begin{enumerate}
    \item To the best of our knowledge, we provide the first comprehensive review of the grid-centric perception methods from various perspectives for autonomous driving.
    \item We provide a structural and hierarchical overview of grid-centric perception techniques. Both academic and industry perspectives on the grid-centric perception of autonomous driving practices are analyzed.
    \item We summarize the observation of the current trend and provide future outlooks for grid-centric perceptions.
\end{enumerate}

\new{
As illustrated in Fig. \ref{fig:taxonomy}, this paper is organized in a hierarchically-structured taxonomy. We focus on four core issues in the taxonomy, the spatial representation of features, the temporal expression of features, efficient learning, and planning on grid-centric perception. Section. \ref{background} introduces the background of grid-centric perception, including task introduction and comparison with object-centric pipelines, commonly used datasets, popular benchmarks and metrics. Section. \ref{sec:3d_occu} discusses data preparation, network design and deployment techniques for 3D occupancy networks. Section. \ref{sec:temporal} introduces temporal fusion, motion prediction and 4D occupancy forecasting. Section \ref{sec:effilearn} introduces label-efficient paradigm including learning open-vocabulary occupancy with self-supervised learning and using occupancy as unified pretraining for downstream tasks. Section. \ref{sec:driveapplication} introduces how planning methods benefit from grid-centric perception as input. Section. \ref{Discussion} discusses limitations of the current trend and presents future outlooks for grid-centric perception techniques. Section. \ref{conclu} concludes this paper. 
}

\new{
Despite introduction of development in grid-centric perception in the past five years, a review of well-established non-deep-learning OGM and its variants is included in the Appendix. Moreover, general BEV perception pipelines are also discussed in the Appendix.
}
\begin{figure*}[thpb]
	\centering
	\includegraphics[width=1\textwidth]{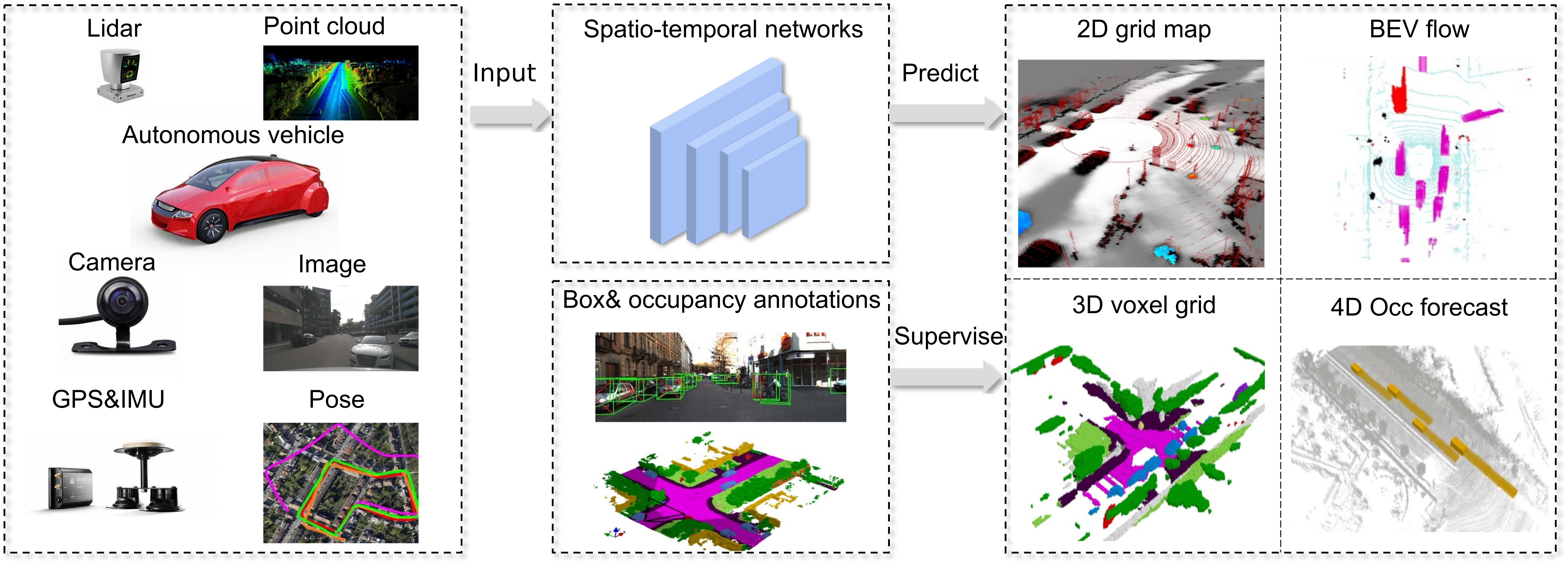}
	\caption{An illustration of grid-centric perception in autonomous driving scenarios. (Left): The autonomous vehicle is equipped with different sensors such as LiDARs, radars and cameras, as well as GPS and IMU for localization. (Middle): The raw data, point clouds and images are processed in spatio-temporal networks with the supervision from annotations of bounding boxes or occupancy. (Right): Different perception outputs for grid-centric perception. Images source from \cite{MonoScene,MT-DOGM,cam4docc,MotionNet}.}
	\label{intro}
\end{figure*}

% A regularly updated project page is also provided at {\color{blue} \url{https://github.com/synsin0/Grid-Centric-Perception}}.

\section{Background}\label{background}
This section introduces general task formulation, common datasets and popular benchmarks and metrics of grid-centric perception.

\subsection{Task Introduction}
Grid-centric perception refers to the concept that, given the multi-modality input of on-board sensors, algorithms need to transform raw information to BEV or voxel grids and preform various perception tasks on each grid cells. A general formula of grid-centric perception can be represented as:
\begin{equation}
    \mathcal{G} = f(I_{sensor}),
\end{equation}
where $\mathcal{G}$ is a set of the past and future grid-level representation, and $I_{sensor}$ represents one or more sensory inputs. How to represent grid attributes and grid features are two crucial problems in this task. An illustration of grid-centric perception process is shown in Fig. \ref{intro}. Depending on the task and annotation, occupancy expression comes in a variety of forms, including 2D, 3D, and 4D semantic occupancy and flow prediction.

\new{
\textbf{Analysis: Comparisons with other perception pipelines.} Object-centric pipelines, including 3D object detection, tracking and prediction, focuses on representing common-sized well-defined obstacles using 3D bounding boxes. However, novel classes and instances are inevitable on an open street. There are several solutions to this long-tailed problem. In image domain, open-world\cite{OWOD,OW-DETR,PROB} detection methods learns an objectiveness learner to learn where is an object without knowing the specific category. In point cloud domain, Openset3D\cite{OpenSet3D} adopts a similar open-world paradigm. This approach has limitation that unknown objects must look like objects, with no severe occlusion and kinds of similarity with known objects. 
}

\new{
Segmentation-based pipelines, including image segmentation and point cloud segmentation, segment unknown objects from raw sensor data. Open-world image segmentation uses large vision models(e.g. Segment-Anything\cite{SAM}). Open-vocabulary point cloud segmentation relies on distillation across calibrated point cloud and image models. The problem with the segmentation-based method is that it is difficult to overcome the noise of the sensor data itself, for example, the problem of "ghosting" and "expansion" of the high-intensity LiDAR point cloud, and the problem of shooting under low light and back-light in vision. Besides, multi-sensor fusion strategies are usually complicated for segmentation-based pipelines.
}

\new{
Grid-centric pipelines segment low-level occupancy and semantic cues for road obstacles. Grid-centric perception has several advantages: It loosens the restriction on the shape of obstacles and can describe articulated objects with variable shapes; it relaxes the typicality requirements of obstacles. It can accurately describe occupancy and motion cues for novel classes and instances, thus enhancing the system's robustness. Furthermore, the noise of raw sensor data can be eliminated with model designs or multi-modal fusion. 
}

\new{
In essence, occupancy perception follows the spatial perception first, semantics perception second philosophy. The system first understands spatial occupancy, clusters foreground objects, and then recognizes semantics. Since autonomous driving is spatially sensitive, but does not require high accuracy for the exact semantic categories of obstacles. As a typical example, MotionNet\cite{MotionNet} can successfully detect the disabled with neither learning the specific categories nor classifying them as pedestrian. occupancy perception is well suited for robust open-world 3D perception.
}
\subsection{Datasets}
Grid-centric methods are mostly conducted on existing large-scale autonomous driving datasets with annotations of 3D object bounding boxes, LiDAR segmentation labels, annotations of 2D\&3D lane and high definition maps. Moreover, new benchmarks and 3D occupancy labels are built upon existing datasets.These datasets include KITTI\cite{KITTI}, nuScenes\cite{nuScenes}, Argoverse\cite{Argoverse}, Lyft L5\cite{Lyft}, SemanticKITTI\cite{SemanticKITTI}, KITTI-360\cite{KITTI-360}, Waymo Open Dataset (WOD)\cite{Waymo}, Argoverse2\cite{Argoverse2} and ONCE dataset\cite{Once}. Table. \ref{dataset_comparison} provides a summary of these benchmarks' information.

\subsection{Benchmarks and Evaluation Metrics}

This section introduces occupancy-oriented benchmarks built on top of public datasets and their evaluation metrics.

\subsubsection{BEV Segmentation Benchmarks} 

BEV segmentation is defined as semantic or instance segmentation of BEV grids. The common categories include dynamic objects (vehicles, trucks, pedestrians and cyclists), static road layouts and map elements (lanes, pedestrian crossing, drivable areas, walkway). 

\textbf{Metrics for BEV segmentation.} 
For free-space segmentation, which classifies grids as occupied and free, most works use accuracy for a simple metric. For semantic segmentation, the primary metric is \textbf{Intersection-over-Union} (IoU) for each class and mean Intersection over Union (mIoU) over all classes. IoU between prediction $A$ and label $B$ writes: 
\begin{equation}
	\mathrm{IoU}(A,B)=\frac{A\cap B }{A\cup B},
\end{equation}

\subsubsection{3D Occupancy Prediction Benchmarks} 

\textbf{Semantic scene completion.}
SemanticKITTI\cite{SemanticKITTI} dataset first defines the task of outdoor semantic scene completion (SSC). The SSC task is original designed to predict the complete scene inside a certain volume, given single-scan LiDAR pointclouds, but it is further extended to cameras. In the scene around the ego vehicle, the volume is represented by uniform voxel grids, each of which possesses the property of occupancy (empty or occupied) and its semantic label. SSCBench\cite{SSCBench} annotates SSC labels on KITTI-360\cite{KITTI-360}, nuScenes\cite{nuScenes} and Waymo open dataset\cite{Waymo}.

\new{
\textbf{3D occupancy prediction (Occ3D).}
After Tesla's report of its occupancy network, 3D occupancy benchmarks \cite{Occ3D,OpenOccupancy} are built upon public datasets. 3D occupancy prediction is formalized as jointly estimating the occupancy state and semantic label of each voxel in the scene. The occupancy state is divided into free, occupied or unobserved, and the semantic label includes predefined categories and general objects. 
}

\new{
Commonly-used occupancy benchmarks include Occ3D-nuScenes\cite{Occ3D}, OpenOccupancy-nuScenes\cite{OpenOccupancy} and Occ3D-Waymo\cite{Occ3D}. Recent challenge is held to promote the research of vision-centric 3D occupancy prediction, including 3D occupancy prediction at CVPR 2023 Workshops\footnote{\url{https://opendrivelab.com/AD23Challenge.html##Track3}} based on Occ3D-nuScenes \cite{Occ3D}.
}

\new{
\textbf{Analysis: Differences in task formulations of SSC and Occ3D.} There are two major differences between SSC and Occ3D. One is occlusion handling: SSC completes occluded areas based on visible areas, while Occ3D only evaluates on visible regions. The other is dynamic objects, SSC assumes the scene is static in ground-truth generation while Occ3D predicts dynamic objects.  
}

\new{
\textbf{Metrics for 3D occupancy prediction (Occ3D) and semantic scene completion (SSC).} 
The primary metrics for semantic scene completion is \textbf{mIoU} over all semantic classes. \textbf{IoU}, Precision and Recall are used on the scene completion to assess the geometrical reconstruction quality. 3D occupancy prediction challenge measures the F-score as the harmonic mean of the completeness $P_c$ and the accuracy $P_a$, F-score is computed as follows:
\begin{equation}
    F-score = (\frac{P_c^{-1}+P_a^{-1}}{2})^{-1}
\end{equation}
}

\new{
where $P_a$ is the percentage of predicted voxels that are within a distance threshold to the ground truth voxels, and $P_c$ is the percentage of ground truth voxels that are within a distance threshold to the predicted voxels. All metrics are only evaluated in annotated space due to semi-dense ground truth in most real-world datasets.
}

\subsubsection{Occupancy and Flow Prediction Benchmarks}
In addition to occupancy, velocity information is crucial to the safety of dynamic scenes. Traditional object-centric pipeline gets velocity in tracking, which reduces reliability. In contrast, obtaining the occupancy velocity directly on spatial-temporal grids can bypass the matching process in training and obtain a more robust motion estimation.

\textbf{BEV occupancy motion and flow.}
BEV occupancy tasks have different formulations. MotionNet defines the BEV Motion task as to predict the short-term future motion displacement of each grid cell. That is, how far each grid cell may move in the next second. FIERY\cite{FIERY} formulates the flow task as future video instance segmentation tasks in the next two seconds. Waymo introduces long-term (to 8s future) occupancy prediction while serves as a useful supplementary of trajectory-set prediction and holds an occupancy and flow challenge at the CVPR2022 workshop\footnote{\url{https://waymo.com/open/challenges/2022/occupancy-flow-prediction-challenge/}} The task stipulates that, given a one-second history of real agents in a scene, the task must predict the flow fields of all agents in eight seconds.

\new{
\textbf{3D occupancy and flow.}
OpenOCC \cite{OpenOcc} considers the foreground object motion with flow (velocity) annotation of object voxels to form a new 3D occupancy benchmark based on nuScenes \cite{nuScenes}. CVPR 2024 workshop holds Occupancy and Flow Challenge\footnote{\url{https://opendrivelab.com/challenge2024/\#occupancy_and_flow}}, which requires predicting the semantics and flow of each voxel in the scene given images from multiple cameras.
}

\new{
\textbf{Analysis: Comparisons with scene flow.} optical flow and scene flow\cite{PointMotionNet} are well-established to estimate the motion of image pixels or LiDAR points of adjacent frames. In contrast, BEV flow are more structured representation of dynamic scenes, which runs much faster than disorganized pointclouds and dense image pixels. Moreover, scene flow focuses on instantaneous movement while BEV flow extends to farther future occupancy.
}

\begin{table*}[thpb]
\renewcommand\arraystretch{1.5}

\resizebox{\textwidth}{!}{
	\begin{tabular}{cccccccccccccc}
		\hline
		Dataset       & Year & Size(hr.) & Scenes & LiDAR scans & RGB images & Ann. frames & Annotations & Classes & Night/Rain & Views & Stereo & Locations & Auxiliary      \\ \hline
		KITTI\cite{KITTI}         & 2012 & 1.5       & 22     & 15k         & 15k        & 15k         & 200k        & 8       & No/No      & 1     & Yes    & Germany   & -               \\
		Lyft L5\cite{Lyft}       & 2019 & 2.5       & 366    & 46k         & 323k       & 46k         & 1.3M        & 9       & No/No      & 6     & No     & USA       & Maps            \\
		Argoverse\cite{Argoverse}     & 2019 & 0.6       & 113    & 44k         & 490k       & 22k         & 993k        & 15      & Yes/Yes    & 7     & Yes    & USA       & Maps            \\
		SemanticKITTI*\cite{SemanticKITTI} & 2019 & -         & 22     & 43k         & -          & -           & 4549M*       & 28      & -/-        & -     & -      & Germany   & -               \\
		nuScenes\cite{nuScenes}      & 2020 & 5.5       & 1000   & 400k        & 1.4M       & 40k         & 1,4M        & 23      & Yes/Yes    & 6     & No     & SG,USA    & Maps,RADAR data \\
		Waymo Open\cite{Waymo}    & 2020 & 6.4       & 1150   & 230k        & 1M         & 230k        & 12M         & 23(4)**      & Yes/Yes    & 5     & No     & USA       & -               \\
		KITTI-360\cite{KITTI-360}     & 2021 & -         & -      & 80k         & 300k       & -           & 68k         & 37      & -/-        & 3     & Yes    & Germany   & -               \\
		Argoverse 2\cite{Argoverse2}   & 2021 & -         & 1000   & -           & -          & -           & -           & 30      & Yes/Yes    & 9     & Yes    & USA       & Maps            \\
		ONCE\cite{Once}          & 2021 & 144       & -      & 1M          & 7M         & -           & 417k        & 5       & Yes/Yes    & 7     & No     & China     & -               \\ \hline
	\end{tabular}}
	\label{dataset_comparison}
\caption{Detailed information of influential datasets for grid-centric perception. (*) The KITTI-based SemanticKITTI Dataset contains LiDAR scans with dense point-wise annotations. For additional dataset annotations, we indicate the number of bounding boxes. (**) Waymo Open Dataset provides dense labels for each LiDAR point with 23 classes and 3D bounding box labels with 4 classes. (-) indicates that no information is provided or the item does not exist in this dataset. SG: Singapore.}
\end{table*}

% Future development of datasets for grid-centric perception. Current driving datasets are mostly intended for benchmarking fully-supervised closed-world object-centric tasks, which may impede the unique advantages of grid-centric perception. Future datasets may require a more varied open-world driving situation in which potential obstacles cannot be represented as bounding boxes. Argoverse2\cite{Argoverse2} dataset is a next-generation dataset for its 10Hz densely-annotated 1k sensor sequences with 26 categories and super large-scale, unlabeled 6M LiDAR frames.

\textbf{Metrics for BEV occupancy and flow prediction.}
MotionNet\cite{MotionNet} encodes motion information by associating each grid cell with a displacement vector in a BEV map and proposes metrics for motion prediction by classifying non-empty grid cells into three velocity ranges: static, slow($\le5\mathrm{m/s}$) and fast($>5\mathrm{m/s}$). In each velocity range, the mean and median $L_2$ distances between the predicted displacements and the ground-truth displacements have been utilized.

FIERY\cite{FIERY} uses the \textbf{Video Panoptic Quality} (VPQ)\cite{VPQ} metric for prediction of future instance segmentation and motion in a BEV map. This metric is defined as:
\begin{equation}
	\mathrm{VPQ}=\sum_{t=0}^{H} \frac{ {\textstyle \sum_{(x_{t},y_{t})\in TP_{t}}\mathrm{IoU}(x_{t},y_{t})} }{\left | TP_{t} \right |+\frac{1}{2} \left | FP_{t} \right |+\frac{1}{2} \left | FN_{t} \right | },
\end{equation}
where $TP_{t}$, $FP_{t}$, $FN_{t}$ are the true positive, false positive and false negative instance predictions at timestep $t$, $H$ is the future predition horizon. Notice that a true positive prediction has IoU over 0.5 with the ground truth and the same instance id as the ground truth throughout time.

Waymo's Occupancy Flow Challenge evaluates the ability of algorithms to predict occupancy distribution across a longer time span (8s). Metrics of occupancy flow consist of occupancy metrics, flow metric, and joint metrics. The primary occupancy metrics are \textbf{Area under the Curve} (AUC) and \textbf{Soft Intersection over Union} (Soft-IoU) \cite{Soft-IoU}which are used for binary segmentation. AUC$(O_{t}^{\mathcal{K}},\hat{O}_{t}^{\mathcal{K} }  )$ for class $\mathcal{K}$ (vehicle/pedestrian) estimates the area under the PR-curve. The Soft-IoU metric for each class $\mathcal{K}$ writes:
\begin{equation}
	\mathrm{Soft\text{-}IoU}(O_{t}^{\mathcal{K}},\hat{O}_{t}^{\mathcal{K}})=\frac{ {\textstyle \sum_{x,y}} O_{t}^{\mathcal{K}}\cdot \hat{O}_{t}^{\mathcal{K}}}{ {\textstyle \sum_{x,y}O_{t}^{\mathcal{K}}+ \hat{O}_{t}^{\mathcal{K}}-O_{t}^{\mathcal{K}}\cdot \hat{O}_{t}^{\mathcal{K}}}}, 
\end{equation}
where $O_t,\hat{O}_t$ are ground-truth and predicted occupancy at time step $t$ .

For flow metric, \textbf{End-Point Error} (EPE) measures the mean L2 distance between the ground-truth flow field $F_{t}^{\mathcal{K}}(x,y)$ and predicted flow field $\hat{F}_{t}^{\mathcal{K}}(x,y)$ as:
\begin{equation}
	\left \| F_{t}^{\mathcal{K}}(x,y)-\hat{F}_{t}^{\mathcal{K}}(x,y)\right \|_{2}, where\; F_{t}^{\mathcal{K}}(x,y) \neq (0,0),
\end{equation}
where a flow field $F_{t}$ at time t holds motion vector $(dx, dy)$ for each pixel.

The joint metrics measure the accuracy of flow and occupancy predictions at each time step $t$, so $\hat{F}_{t}$ is used to warp the ground-truth occupancy ($O_{t-1}$) as:
\begin{equation}
	\hat{W}_{t}=\hat{F}_{t}\circ O_{t-1},
\end{equation}
where $\circ$ applies the flow field as a function to transform the occupancy.
If the joint prediction is accurate enough, $\hat{W}_{t}\hat{O}_{t}$ should be close to the ground-truth $O_{t}$. Therefore, \textbf{flow-grounded AUC}$(O_{t}^{\mathcal{K}},\hat{W}_{t}\hat{O}_{t}^{\mathcal{K} }  )$ and \textbf{flow-grounded Soft-IoU}$(O_{t}^{\mathcal{K}},\hat{W}_{t}\hat{O}_{t}^{\mathcal{K} }  )$ have been employed.

\new{
\textbf{Metrics for 3D occupancy and flow challenge.} CVPR 2024 Challenge Occupancy and Flow sets Occupancy Score as the primary metric, which consists of Ray-based mIoU and absolute velocity error (AVE) for occupancy flow. Ray-based mIoU has advantages over per-voxel IoU\cite{SparseOcc_opendrivelab} in that per-voxel IoU has inconsistent penalties of depth estimation and it avoids evaluation of areas unscanned by LiDAR points without the use of the visibility mask.
}

\new{
For ray-based mIoU, query rays are projected into the predicted and the ground-truth occupancy. For each ray, the depth, class label and flow are obtained when it intersects any surface. If the class labels are consistent and the L1 error between the ground-truth depth and the predicted depth is less than a certain threshold, the query ray is classified as a true positive. Similarly, the number of false positive and false negative can be obtained to compute mIoU across classes. Finally, ray-based mIoU is the average under the distance threshold ${1,2,4}$ meters. For absolute velocity error (AVE), the challenge measures velocity errors for true positives under the threshold of $2m$. The final occupancy score writes:
\begin{equation}
    Score_{occ}=0.9\times \mathrm{mIoU}+0.1\times max(1-\mathrm{mAVE},0.0)
\end{equation}
}

\new{
\subsubsection{4D Occupancy Forcasting Benchmarks}
4D Occupancy Forcasting extends 3D occupancy prediction and flow prediction to 4D spatio-temporal occupancy prediction. Given past and current consecutive point clouds\cite{OCFBench} or camera images \cite{cam4docc}, 4D occupancy forecasting aims to output the current occupancy and the occupancy in a future interval in the present coordinate system. 4D occupancy forecasting can be formulated as either supervised or unsupervised and evaluated either with ground-truth labels or raw point clouds sequences. OCFBench \cite{OCFBench} and Cam4DOcc \cite{cam4docc} constructs 4D groundtruth with object tracking labels and 3D occupancy labels, while Argoverse 2 4D Occupancy Challenge\footnote{\url{https://eval.ai/web/challenges/challenge-page/1977/overview}} and CVPR 2024 Predictive World Model Challenge\footnote{\url{https://opendrivelab.com/challenge2024/\#predictive_world_model}} eval the 4D occupancy with ray cast from future LiDAR origin to the future LiDAR point clouds.
}

\new{
\textbf{Metrics for 4d occupancy forecasting.}
If evaluations are with 4D occupancy labels, these benchmarks adopt widely-used metrics, such as mIoU, mAP, precision, recall and F1 score, to assess the model performance along the sequence.
}

\new{
If evaluations are without 4D occupancy labels, every LiDAR point forms a ray from LiDAR origin to the projected surface at a certain direction. The predicted depth is to the distance of marching the rays to the nearest occupied surface at the certain direction. Then the predicted depth are transformed to a predicted point cloud. The Chamfer distance (CD) between predicted and ground-truth point clouds is the primary metrics. L1 error and relative absolute L1 error are also used for evaluation. 4d-occ-forecasting\cite{4d-occ-forecasting} evaluates all future frames together and ViDAR\cite{ViDAR} reports Chamfer disntace per frame. These evaluation usually sets a region of interest (RoI) that are within 4D occupancy grid range.
}

\section{3D Occupancy Mapping}\label{sec:3d_occu}

The field of robotics has a long tradition of navigating with 2D grids. However, driving with BEV grids is prone to failure in some complex traffic scenes, such as the ground is not flat or the obstacle is not uniform in height dimension (suspended object). In contrast, 3D grids are able to represent the full geometry of driving scenes, including the road surface and the shape of obstacles, at the expense of higher computation costs. Therefore, computationally-efficient occupancy network is an important issue for grid-centric perception. 

\new{
In this section, we first review the full process of occupancy networks, how data and labels are prepared, how networks from different modalities are designed and how occupancy networks are deployed on-board. We collate most methods of which the performance are demonstrated on three popular datasets with benchmarks (SSC-SemanticKITTI, Occ3D-nuScenes, OpenOccupancy-nuScenes). The comparisons of typical are available at Table.\ref{performance_SSC}, Table.\ref{performance_nuscenes_occ3d}, Table.\ref{performance_nuscenes_openoccupancy}, respectively.
}
% Semantic scene completion (SSC) and 3D Occupancy prediciton are both formulated as tasks of explicitly inferring the occupancy and semantics of uniform-sized voxels. 

% Compared with the past SSC survey\cite{SSC-Survey} which thoroughly investigates both indoor and outdoor SSC datasets and methods, We focus on advances in occupancy networks for autonomous driving. 

% Due to the above requirements, it is necessary to review the key designs of 3D occupancy network, which are divided into a point cloud centric pipelines and a vision-centric according to sensor modalities. 

\begin{table*}[t!]
		\footnotesize
		\setlength{\tabcolsep}{0.0020\linewidth}

		\newcommand{\classfreq}[1]{{~\tiny(\semkitfreq{#1}\%)}}  %
		\centering
              \scalebox{1}{
              \setlength{\tabcolsep}{0.35mm}{
		\begin{tabular}{l|c|cc| c c c c c c c c c c c c c c c c c c c}
			\toprule
			Method
                & \makecell[c]{Mod.}
                & IoU 
			 & mIoU 
			& \rotatebox{90}{Road}
			& \rotatebox{90}{Sidewalk}
			& \rotatebox{90}{Parking}
			& \rotatebox{90}{Other-grnd} 
			& \rotatebox{90}{Building} 
			& \rotatebox{90}{Car}
			& \rotatebox{90}{Truck}
			& \rotatebox{90}{Bicycle}
			& \rotatebox{90}{Motorcycle} 
			& \rotatebox{90}{Other-veh.}
			& \rotatebox{90}{Vegetation} 
			& \rotatebox{90}{Trunk}
			& \rotatebox{90}{Terrain} 
			& \rotatebox{90}{Person}
			& \rotatebox{90}{Bicyclist}
			& \rotatebox{90}{Motorcyclist.} 
			& \rotatebox{90}{Fence} 
			& \rotatebox{90}{Pole}
			& \rotatebox{90}{Traf.-sign} 
			\\
			\midrule

            S3CNet \cite{S3CNet} &L	&45.60 	&\bf 29.53	&42.00               	&22.50	&17.00	&7.90	&\bf52.20	&31.20	&6.70	&\bf41.50	&\bf45.00	&\bf16.10	&39.50	&\bf34.00	&21.20	&\bf45.90	&\bf35.80	&\bf16.00	&\bf31.30	&\bf31.00	&\bf24.30  \\
            LMSCNet \cite{LMSCNet}	&L &56.72	&17.62	&64.80 	&34.68 	&29.02	&4.62	&38.08	&30.89 	&1.47	&0.00	&0.00	&0.81	&41.31	&19.89 	&32.05	&0.00	&0.00	&0.00	&21.32	&15.01	&0.84 \\
            JS3C-Net \cite{JS3C-Net}	&L &56.60	&23.75	&64.70 	&39.90 	&34.90 	&\bf14.10 	&39.40 	&33.30 	&\bf7.20 	&14.40 	&8.80	&12.70 	&43.10 	&19.60 	&40.50	&8.00 	&5.10 	&0.40 	&30.40 	&18.90 	&15.90 \\
            Local-DIFs \cite{Local-DIFs}	&L &\bf 58.90	&23.56	&\bf69.60 	&\bf44.50	&\bf41.80 	&12.70 	&41.30	&\bf35.40 	&4.70	&3.60 	&2.70	&4.70	&\bf43.80 	&27.40 	&\bf40.90	&2.40	&1.00	&0.00	&30.50	&22.10 	&18.50 \\
            \midrule
            OpenOccupancy \cite{OpenOccupancy}	&C$+$L &-&20.42 	&60.60 	&36.10	&29.00 	&\bf13.00 	&\bf38.40 	&33.80 	&4.70	&3.00 	&2.20	&5.90	&\bf41.50	&20.50 	&35.10	&0.80 	&2.30 	&\bf0.60 	&26.00 	&18.70 	&15.70  \\
            Co-Occ \cite{Co-Occ} &C$+$L &-	&\bf 24.44 	&\bf72.00 	&\bf43.50 	&\bf42.50	&10.20	&35.10 	&\bf40.00 	&\bf6.40 	&\bf4.40 	&\bf3.30	&\bf8.80 	&41.20 	&\bf30.80 	&\bf40.80 	&\bf1.60 	&\bf3.30 	&0.40 	&\bf32.70	&\bf26.60	&\bf20.70 \\
            \midrule
            MonoScene \cite{MonoScene}	&C &34.16	&11.08	&54.70 	&27.10 	&24.80 	&5.70 	&14.40 	&18.80 	&3.30 	&0.50 	&0.70 	&4.40 	&14.90 	&2.40 	&19.50 	&1.00 	&1.40 	&0.40 	&11.10 	&3.30 	&2.10  \\
            TPVFormer \cite{tpvformer}	&C &34.25 	&11.26	&55.10 	&27.20 	&27.40 	&6.50 	&14.80 	&19.20 	&3.70 	&1.00 	&0.50 	&2.30 	&13.90	&2.60 	&20.40 	&1.10 	&2.40 	&0.30 	&11.00 	&2.90 	&1.50 \\
            OccFormer \cite{occformer}	&C &34.53 	&12.32	&55.90 	&30.30 	&\bf31.50 	&6.50 	&15.70 	&21.60 	&1.20 	&1.50 	&1.70 	&3.20 	&16.80 	&3.90 	&21.30 	&2.20 	&1.10 	&0.20 	&11.90 	&3.80 	&3.70  \\
            SurroundOcc \cite{SurroundOcc}	&C &34.72 	&11.86	&56.90 	&28.30 	&30.20 	&6.80 	&15.20 	&20.60 	&1.40 	&1.60 	&1.20 	&4.40 	&14.90 	&3.40 	&19.30 	&1.40 	&2.00 	&0.10 	&11.30 	&3.90 	&2.40  \\
            NDC-Scene \cite{Ndc-scene}	&C &36.19	&12.58	&58.12 	&28.05 	&25.31 	&6.53 	&14.90 	&19.13 	&4.77 	&1.93 	&2.07 	&6.69 	&17.94 	&3.49 	&25.01 	&\bf3.44 	&2.77 	&1.64 	&12.85 	&4.43 	&2.96   \\
            RenderOcc \cite{RenderOcc}	&C & -	&8.24	&43.64	&19.10	&12.54	&0.00	&11.59	&14.83	&2.47	&0.42	&0.17	&1.78	&17.61	&1.48	&20.01	&0.94	&3.20	&0.00	&4.71	&1.17	&0.88  \\
            Symphonies \cite{Symphonies}	&C &42.19 	&15.04	&58.40 	&29.30	&26.90 	&\bf11.70 	&\bf24.70 	&23.60 	&3.20	&3.60 	&\bf2.60 	&5.60 	&24.20 	&\bf10.00 	&23.10 	&3.20 	&1.90 	&\bf2.00 	&16.10 	&7.70 	&8.00   \\
            HASSC \cite{HASSC} &C &42.87	&14.38	&55.30	&29.60	&25.90	&11.30	&23.10	&23.00	&\bf9.80	&1.90	&1.50	&4.90	&24.80	&9.80	&26.50	&1.40	&3.00	&0.00	&14.30	&7.00 	&7.10  \\
            BRGScene \cite{BRGScene}	&C &43.34	&15.36	&\bf61.90	&\bf31.20	&30.70	&10.70	&24.20	&22.80	&8.40	&3.40	&2.40	&6.10	&23.80	&8.40	&27.00	&2.90	&2.20	&0.50	&\bf16.50	&7.00	&7.20  \\
            VoxFormer \cite{voxformer}	&C &\bf 44.15	&13.35	&53.57	&26.52	&19.69	&0.42	&19.54	&\bf26.54	&7.26	&1.28	&0.56	&\bf7.81	&\bf26.10	&6.10	&\bf33.06	&1.93	&1.97	&0.00	&7.31	&\bf9.15	&4.94   \\
            MonoOcc \cite{monoocc}	&C &- 	&\bf 15.63	&59.10 	&30.90 	&27.10 	&9.80 	&22.90 	&23.90	&7.20	&\bf4.50 	&2.40	&7.70	&25.00	&9.80	&26.10	&2.80 	&\bf4.70	&0.60 	&16.90 	&7.30 	&\bf8.40   \\

			\bottomrule
		\end{tabular}
  }
  }
		\label{performance_SSC}
          \new{
  		\caption{3D occupancy prediction comparison on the SemanticKITTI test set \cite{SemanticKITTI}. Mod.: Modality. C: Camera. L: LiDAR.
         }
         }
	\end{table*}

\begin{table*}[ht]
        \centering
	\footnotesize
	\setlength\tabcolsep{1.6pt}
	\renewcommand\arraystretch{1.0}
	% \vspace{-10pt}
	% \newcommand{\classfreq}[1]{{~\tiny(\nuscenesfreq{#1}\%)}}  %
 
	% \begin{center}
	% 	\resizebox{\textwidth}{!}{
			\begin{tabular}{l|c|c| c c c c c c c c c c c c c c c c c}
				\toprule
				Method
				& \makecell{Image \\ Backbone} & mIoU
				& \rotatebox{90}{Others}
				
				& \rotatebox{90}{Barrier}
				
				& \rotatebox{90}{Bicycle}
				
				& \rotatebox{90}{Bus}
				
				& \rotatebox{90}{Car}
				
				& \rotatebox{90}{Const. veh.}
				
				& \rotatebox{90}{Motorcycle}
				
				& \rotatebox{90}{Pedestrian}
				
				& \rotatebox{90}{Traffic cone}
				
				& \rotatebox{90}{Trailer}
				
				& \rotatebox{90}{Truck}
				
				& \rotatebox{90}{Drive. suf.}
				
				& \rotatebox{90}{Flat}
				
				& \rotatebox{90}{Sidewalk}
				
				& \rotatebox{90}{Terrain}
				
				& \rotatebox{90}{Manmade}
				
				& \rotatebox{90}{Vegetation}
				
				\\
				\midrule
				\multicolumn{20}{c}{Performances on Validation Set} \\
				\midrule
				MonoScene\cite{MonoScene}  & R101-DCN & 6.06 & 1.75 & 7.23 & 4.26 & 4.93 & 9.38 & 5.67 & 3.98 & 3.01 & 5.90 & 4.45 & 7.17 & 14.91 & 6.32 & 7.92 & 7.43 & 1.01 & 7.65\\

				%			BEVDet  & R101-DCN & 11.73 & 2.09 & 15.29 & 0.00 & 4.18 & 12.97 & 1.35 & 0.00 & 0.43 & 0.13 & 6.59 & 6.66 & 52.72 & 19.04 & 26.45 & 21.78 & 14.51 & 15.26\\
				%			BEVFormer  & R101-DCN & 26.88 & 5.85 & 37.83 & 17.87 & 40.44 & 42.43 & 7.36 & 23.88 & 21.81 & 20.98 & 22.38 & 30.70 & 55.35 & 28.36 & 36.0 & 28.06 & 20.04 & 17.69 \\
				CTF-Occ\cite{Occ3D}  & R101-DCN & 28.53 & 8.09 & 39.33 & 20.56 & 38.29 & 42.24 & 16.93 & 24.52 & 22.72 & 21.05 & 22.98 & 31.11 & 53.33 & 33.84 & 37.98 & 33.23 & 20.79 & 18.00\\
				BEVFormer\cite{BEVFormer}  & R101-DCN & 39.24 & 10.13 & 47.91 & 24.90 & 47.57 & 54.52 & 20.23 & 28.85 & 28.02 & 25.73 & 33.03 & 38.56 & 81.98 & 40.65 & 50.93 & 53.02 & 43.86& 37.15  \\
				PanoOcc\cite{PanoOcc}  & R101-DCN & {42.13} & 11.67 & 50.48 & 29.64 & 49.44 & 55.52 & 23.29 & \textbf{33.26} & 30.55 & 30.99 & 34.43 & 42.57 & \textbf{83.31} & 44.23 & 54.40 & 56.04 & 45.94 & 40.40  \\
				BEVDet\cite{BEVDet}  & Swin-B & 42.02 & 12.15 & 49.63 & 25.10 & 52.02 & 54.46 & 27.87 & 27.99 & 28.94 & 27.23 & 36.43 & 42.22 & 82.31 & 43.29 & 54.62 & 57.90 & 48.61 & 43.55  \\ 
				
				% \midrule
				% Baseline (ours) & Swin-B & 43.73 & \textbf{12.66}& 51.83 & 31.14 & 51.67 & 56.38 & 30.03 & 32.38 &31.27 &30.94 &39.06 &44.22 &82.79 &44.56 &55.07 & \textbf{58.51} &48.11&42.71\\
                % Baseline (ours) & Swin-B & 44.14 & \textbf{13.39}& 52.20 & \textbf{31.43} & 52.01 & 56.70 & \textbf{30.66} & 32.95 &31.56 &\textbf{31.31} &39.87 &44.64 &82.98 &\textbf{44.97} &\textbf{55.43} & \textbf{58.90} &48.43&42.99\\
				RadOcc\cite{RadOcc}  & Swin-B & \textbf{46.06} &9.78 &\textbf{54.93} &20.44 &\textbf{55.24} &\textbf{59.62} &30.48 &28.94 &\textbf{44.66} &28.04 &\textbf{45.69} &\textbf{48.05} &81.41 &39.80 &52.78 &56.16 &\textbf{64.45} &\textbf{62.64}\\
				\midrule

				\multicolumn{20}{c}{Performances on  3D Occupancy Prediction Challenge} \\
				\midrule
				BEVFormer\cite{BEVFormer} & R101-DCN  & 23.70 & 10.24&36.77&11.70&29.87&38.92&10.29&22.05&16.21&14.69&27.44&33.13&48.19&33.10&29.80&17.64&19.01&13.75 \\
				SurroundOcc\cite{SurroundOcc} & R101-DCN & 42.26 &11.70&50.55&32.09&41.59&57.38&27.93&38.08 &30.56&29.32&48.29&38.72&80.21&48.56&53.20&47.56&46.55&36.14 \\
				BEVDet\cite{BEVDet} & Swin-B &42.83&18.66&49.82&31.79&41.90&56.52&26.74&37.31&30.01&31.33&48.18&38.59&80.95&50.59&53.87&49.67&46.62&35.62 \\
				PanoOcc\cite{PanoOcc} & Intern-XL & 47.16 & 23.37 & 50.28 & 36.02 & 47.32 & 59.61 & 31.58 & 39.59 & 34.58 & 33.83 & 52.25 & 43.29 & 83.82 & 55.81 & 59.41 & 53.81 & 53.48 & 43.61 \\
				% \midrule
				% Baseline-T  (ours) & Swin-B & 46.55 &\textbf{24.28}&52.81&34.78&44.27&56.78&31.69&41.01&33.56&36.40&48.20&41.43&{83.81}&{54.96}&58.59&54.04&53.26&41.51\\
				% Baseline-T  (ours) & Swin-B & 47.74 &22.88&50.74&\textbf{41.02}&\textbf{49.39}&55.40&33.41&45.71&38.57&35.79&48.94&44.40&{83.19}&{52.26}&59.09&\textbf{55.83}&51.35&43.54\\
				RadOcc\cite{RadOcc}  & Swin-B & 49.98&21.13&55.17&39.31&48.99&59.92&33.99&46.31&\textbf{43.26}&39.29&52.88&44.85&83.72&53.93&59.17&55.62&60.53&51.55\\
                    UniOcc\cite{UniOcc} & ConvNeXt-Base & 51.27 & 26.94 & 56.17 & 39.55 & 49.40 & 60.42 & 35.51 & 44.77 & 42.96 & 38.45 & 59.33 & 45.90 & 83.90 & 53.53 & 59.45 & 56.58 & 63.82 & \textbf{54.98} \\
                    MiLO\cite{MiLO} & Intern-XL\&Swin-L & 52.45 & 27.80 & 56.28 & 42.62 & 50.27 & 61.01 & 35.41 & 47.97 & 38.90 & 40.29 & 56.66 & 47.03 & 86.96 & 57.48 & 63.64 & 62.53 & 63.00 & 53.74  \\
                    FB-Occ\cite{FB-OCC} & Intern-H & \textbf{54.19} & \textbf{28.95} & \textbf{57.98} & \textbf{46.40} & \textbf{52.36} & \textbf{63.07} & \textbf{35.68} & \textbf{48.81} & 42.98 & \textbf{41.75} & \textbf{60.82} & \textbf{49.56} & \textbf{87.29} & \textbf{58.29} & \textbf{65.93} & \textbf{63.30} & \textbf{64.28} & 53.76 \\ 
				\bottomrule
			\end{tabular}
                
	% 	}
	% \end{center}
	%	\vspace{-.6cm}
         \new{
	   \caption{3D occupancy prediction performance on the Occ3D-nuScenes. All methods only use vision as input. CVPR20223 3D occupancy prediction challenge adopts Occ3D-nuScenes\cite{Occ3D} test set. }
    }
    \label{performance_nuscenes_occ3d}%\vspace
\end{table*}

\begin{table*}[h]
\setlength{\tabcolsep}{0.0045\linewidth}
\newcommand{\classfreq}[1]{{~\tiny(\semkitfreq{#1}\%)}}  %
\centering

\resizebox{1.0\linewidth}{!}{
\begin{tabular}{l|c | c c | c c c c c c c c c c c c c c c c}
    \toprule
    Method
    & \makecell[c]{Modality}
    & IoU& mIoU
    & \rotatebox{90}{Barrier}
    & \rotatebox{90}{Bicycle}
    & \rotatebox{90}{Bus}
    & \rotatebox{90}{Car}
    & \rotatebox{90}{Const. veh.}
    & \rotatebox{90}{Motorcycle}
    & \rotatebox{90}{Pedestrian}
    & \rotatebox{90}{Traffic cone}
    & \rotatebox{90}{Trailer}
    & \rotatebox{90}{Truck}
    & \rotatebox{90}{Drive. suf.}
    & \rotatebox{90}{Other flat}
    & \rotatebox{90}{Sidewalk}
    & \rotatebox{90}{Terrain}
    & \rotatebox{90}{Manmade}
    & \rotatebox{90}{Vegetation} \\
    % & mIoU\\
    \midrule
    MonoScene~\cite{MonoScene} & C   & 18.4 & 6.9 & 7.1  & 3.9  &  9.3 &  7.2 & 5.6  & 3.0  &  5.9& 4.4& 4.9 & 4.2 & 14.9 & 6.3  & 7.9 & 7.4  & 10.0 & 7.6 \\
  
    TPVFormer~\cite{tpvformer} &C &  15.3 &  7.8 & 9.3  & 4.1  &  11.3 &  10.1 & 5.2  & 4.3  & 5.9 & 5.3&  6.8& 6.5 & 13.6 & 9.0  & 8.3 & 8.0  & 9.2 & 8.2 \\

    3DSketch~\cite{3DSketch} &  C\&D & 25.6 & 10.7  & 12.0 &  5.1 &  10.7 &  12.4 & 6.5  & 4.0  & 5.0 & 6.3&  8.0&  7.2& 21.8 &  14.8 & 13.0 &  11.8 & 12.0 & 21.2 \\
        
    AICNet~\cite{AICNet} & C\&D   &   23.8 & 10.6  & 11.5  & 4.0  & 11.8  & 12.3&  5.1 & 3.8  & 6.2  & 6.0 & 8.2&  7.5&  24.1 & 13.0 & 12.8  & 11.5 & 11.6  &  20.2\\

        LMSCNet~\cite{LMSCNet} & L &   27.3 & 11.5 & 12.4&  4.2 & 12.8  & 12.1  & 6.2  &  4.7 & 6.2 & 6.3&  8.8&  7.2& 24.2 & 12.3  & 16.6 & 14.1  & 13.9 & 22.2 \\

        JS3C-Net~\cite{JS3C-Net} &L &   30.2  & 12.5 & 14.2 & 3.4  & 13.6  & 12.0  & 7.2  &  4.3 & 7.3 & 6.8&  9.2& 9.1 & 27.9 & 15.3  & 14.9 & 16.2  & 14.0 & 24.9 \\

        Point-Occ~\cite{PointOcc} & L &\textbf{34.1} &\textbf{23.9} &24.9 &\textbf{19.0} & 20.9& 25.7 &13.4 &\textbf{25.6} &\textbf{30.6} &\textbf{17.9} &16.7 &21.2 &36.5 &\textbf{25.6} &\textbf{25.7} &\textbf{24.9} &\textbf{24.8} & \textbf{29.0} \\
         
        OccFusion~\cite{OccFusion1} & C\&L & 31.1 & 17.0 &15.9 &15.1 &15.8 &18.2 &15.0 &17.8 &17.0 &10.4 &10.5& 15.7 &26.0 &19.4 &19.3 &18.2 &17.0 &21.2 \\
        
        M-baseline~\cite{OpenOccupancy} & C\&L &   29.1 & 15.1 & 14.3 & 12.0 & 15.2 & 14.9 & 13.7 &15.0 & 13.1 &9.0 &10.0 &14.5 &23.2 &17.5 &16.1 &17.2 &15.3 & 19.5  \\ 

        M-CONet~\cite{OpenOccupancy} & C\&L &   29.5 & 20.1 &  23.3  & 13.3& 21.2 & 24.3& \textbf{15.3}  & 15.9& 18.0 & 13.3 & 15.3 & 20.7 & 33.2 & 21.0 & 22.5  & 21.5 &19.6 & 23.2  \\

         Co-Occ~\cite{Co-Occ} & C\&L &30.6&21.9&\textbf{26.5} &16.8&22.3&\textbf{27.0}&10.1&20.9&20.7 &14.5 &16.4 &21.6&\textbf{36.9}&23.5&25.5 &23.7 &20.5 &23.5 \\

         OccGen~\cite{OccGen} & C\&L &30.3&22.0& 24.9&16.4&\textbf{22.5}&26.1 &14.0 &20.1&21.6&14.6&\textbf{17.4}&\textbf{21.9}&35.8&24.5&24.7&24.0&20.5 &23.5 \\
         
    \bottomrule

    \end{tabular}}
    \label{performance_nuscenes_openoccupancy}
    \new{
    \caption{3D semantic occupancy prediction results on nuScenes-OpenOccupancy validation set. We report the geometric metric IoU, semantic metric mIoU, and the IoU for each semantic class. The C, D, and L denotes camera, depth, and LiDAR, respectively.}
    }
\end{table*}

\new{
\subsection{Occupancy Data Pipelines}\label{sec:occ_data_piplie}
\subsubsection{Ground-truth Label Generation}\label{sec:label_gen}
Unlike other perception tasks which directly annotate labels on raw sensor data, occupancy label needs a second-stage annotation generally dependent on the LiDAR segmentation labels for background stuff and bounding boxes for foreground objects. The temporally continuous multi-frame point cloud is transformed to the current position based on ego pose coordinates, and then the point cloud is voxelized with a fixed range and resolution to obtain voxel occupancy and semantic labels. A label generation toolkit example is available at SurroundOcc repository\footnote{\url{https://github.com/weiyithu/SurroundOcc}}. 
}

\new{
Another important label for Occ3D task is the visibility mask for each sensor, as predicting invisible regions is not reasonable for occupancy prediction. Most networks improve slightly when trained with visibility mask. In Occ3D setup, the LiDAR visibility mask is computed with each ray from the origin to the point. a voxel is marked visible if it reflects LiDAR points as occupied or is traversed through by a ray as free; otherwise, the voxel is tagged as inivisible or unobserved region. For camera mask, occulusion is reasoned by forming a virtual ray and sets the nearest occupied voxel as occupied, the traversal as free and voxels behind the nearest observation as invisible. 
}

\new{
In terms of specific operation, the truth effect of Occ3D task and SSC task is slightly different, and there are also their own limitations.
}

\new{
\textbf{Analysis: Limitation of SSC label.} 
Existing outdoor SSC benchmarks\cite{SemanticKITTI,KITTI-360} generate labels by aggregating all-frame semantic pointclouds. The label is dense but traces of dynamic objects are unavoidable interference in labels, dubbed sptaio-temporal tubes in MotionSC\cite{MotionSC}. Due to a large number of parked vehicles in SemanticKITTI, all existing SSC methods predict dynamic objects as if they are static and are punished by benchmark metrics, which is the main reason that SSC tasks are primarily for static scene reconstruction.
}

\new{
\textbf{Analysis: Limitation of Occ3D label.} 
Occ3D bechmarks\cite{Occ3D, OpenOccupancy} generate labels by aggregating the adjacent past and future 20 sweeps from the current frame. Then camera visible mask and lidar visible mask are provided for each frame. This approch can accurately provide labels of dynamic objects but the key drawback is the sparsity of occupied labels, for example, noncontinuous distant ground where only voxels occupied by lidar points are identified as ground and others as empty.
}

\new{
\textbf{Analysis: Future development of label generation.}
The quality of annotation has a great impact on the training effect. However, the cost of the existing secondary annotation method relying on lidar segmentation is very high, which seriously limits the scalability of the occupied network, and on the other hand, the point cloud as a ground-truth system also has the inherent defect of sparsity. In the future, One promising solution is based on neural rendering, using Nerf\cite{NeRF} or Gaussian Splatting\cite{3dgs} to reconstruct the traffic scene in 4D scenes, and then voxelize the neural field to generate a ground-truth value. Emernerf\cite{Emernerf} has made a valuable attempt to obtain a satisfactory occupancy of the truth value in a fully self-supervised manner when the 4d nerf reconstruction quality is compared with Occ3D-Waymo labels.
}

\subsection{Occupancy Networks}\label{sec:occnets}

\subsubsection{LiDAR-based Occupancy Networks}\label{sec:lidarssc}

As LiDAR sensors are naturally suitable for 3D occupancy grids, the challenge of LiDAR-based reconstruction is to infer full scene geometry from points reflected at the surface of obstacles and to infer dense geometry of unobserved areas from sparse LiDAR inputs.

\textbf{LiDAR Occ3D with 2D CNNs.} The dense 3D CNN blocks in SSCNet and TS3D leads to high memory and computation need and dilation of the data manifold. One alternative to address this issue is to take advantage of the efficiency of 2D CNN. LMSCNet\cite{LMSCNet} uses a lightweight U-Net architecture with 2D backbone convolution and 3D segmentation heads. Turning the height dimension into a feature dimension becomes a common practice for traffic scenes where the data mainly varies longitudinally and laterally. Pillar-based LMSCNet achieves good performance at speed and has the capability to infer multiscale SSC. Similarly, Local-DIFs\cite{Local-DIFs} creates a BEV feature map of the point cloud and passes it through 2D U-Net to output feature maps at three scales which make up the novel representation of the 3D scene, continuous Deep Implicit Functions (DIFs). By querying the function for corner points of all voxels, Local-DIFs can be evaluated on SemanticKITTI benchmark and performs well on geometric completion accuracy.

\textbf{LiDAR Occ3D with Sparse Convs.} Another promising alternative is to use sparse 3D networks, such as SparseConv\cite{SparseConv} used in JS3C-Net\cite{JS3C-Net} and Minkowski\cite{Minkowski} used in S3CNet\cite{S3CNet}, which only operate on non-empty voxels. JS3C-Net is a sparse LiDAR point cloud semantic segmentation framework which regards SSC as an auxiliary task. It includes a point-voxel interaction(PVI) module to enhance this multi-task learning and promote knowledge transfer between two tasks. For semantic segmentation, it utilizes a 3D sparse convolution U-Net. The cascaded SSC module predicts a coarse completion result, which is refined in PVI module. Experiments show that JS3C-Net achieves state-of-the-art results on both tasks. S3CNet constructs sparse 2D and 3D feature from a single LiDAR scan and passes them throngh sparse 2D and 3D U-Net style network in parallel. To avoid applying dense convolutions in decoder, S3CNet proposes a dynamic voxel late fusion of BEV and 3D predictions to further densify the scene and then applys a spatial propagation network to refine the result. In particular, it achieves impressive results in rare classes of SemanticKITTI.

\begin{figure}[ht!]
	\centering
	\includegraphics[width=0.45\textwidth]{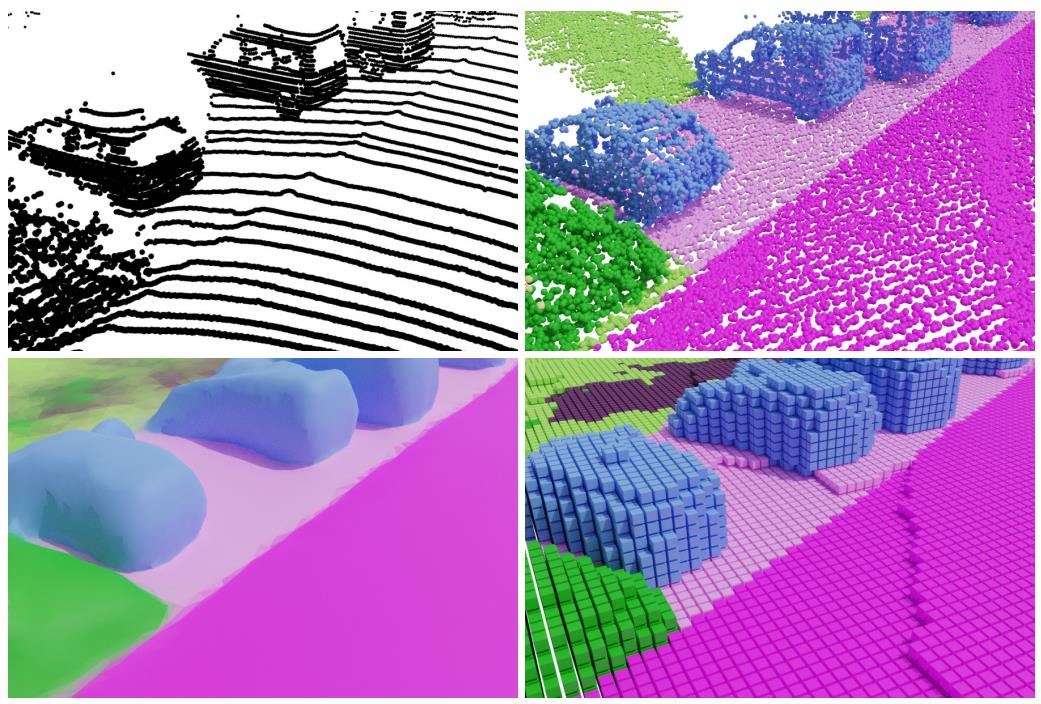}
	\caption{Local-DIF\cite{Local-DIFs}: Local-DIF generates continuous representation for 3D semantic occupancy, which exhibits quantization artefacts on slanted surfaces (e.g. road plane) or edges between objects resulting from a discretization into voxels.}
	\label{localdif_continuous}
\end{figure}

\new{
\textbf{LiDAR Occ3D with Tri-Perspective View.} PointOcc\cite{PointOcc} is an 2D projected-based model without postprocessing. PointOcc proposes a cylindrical tri-perspective view (Cylindrical TPV) as a novel and effective representation of point clouds and can use any 2D backbone for feature extraction without 3D operators. Each point is sptially pooled to three plane and its final feature is summed from three feature plane to get final segmentation results. The TPV representation is able to prediction point cloud segmentation and occupancy segmentation with two simple segmentation head without complex postprocessing.
}

\new{
\textbf{Learning with non-ideal labels.} To address the limitation of SSC label and focus on SSC in the instant of the input, Local-DIFs\cite{Local-DIFs} propose a dataset variant based on SemanticKITTI by only keeping the single instant scan on dynamic objects and removing free space points within the shadows of dynamic objects. Besides, Local-DIF can continuously represent scenes to avoid artifacts caused by discretization, as shown in Fig. \ref{localdif_continuous}. \new{But the segmentation performance of Local-DIFs for fine geometric details needs to be improved, which is attributed to the dominance of the completion task and the ambiguities in ground truth of small object categories.} Wilson et al.\cite{MotionSC} develops a synthetic outdoor dataset CarlaSC without occlusions and traces surrounding the ego vehicle in CARLA\cite{CARLA}. They propose a real-time dense local semantic mapping method, MotionSC\cite{MotionSC} which conbines a spatial-temporal backbone of MotionNet\cite{MotionNet} and the segmentation head of LMSCNet\cite{LMSCNet}. Note that MotionSC which ignores temporal information also performs well on SemanticKITTI benchmark.
}

\new{
\textbf{Occupancy network with Radar point clouds}. Compared with LiDAR, Radar point clouds are relatively sparse, and traditional 3D Radar has no vertical information, making it difficult to perform 3D occupancy. RadarOcc\cite{RadarOcc} is the first attempt to use 4D imaging Radar for 3D occupancy. The limitation of Radar raster reconstruction is to reserve critical raw signals, and the challenges are reducing large data volume, mitigating sidelobes measurements and interpolation-free feature encoding and aggregation.
}
\subsubsection{Camera-based Occupancy Networks}\label{sec:visionssc}

Camera-based methods are emerging in 3D occupancy field. Different from the offline mapping method represented by structure from motion (SFM), online perception projects pixels into three-dimensional space and predict a uniform 3D occupancy grid possibly with semantic categories. \new{The general pipeline for vision occupancy network is to extend BEV feature representation to voxel feature representation. For example, FB-OCC\cite{FB-OCC} is an extension of FB-BEV\cite{FB-BEV} which uses LSS as forward view transformation and BEVFormer as backward view transformation. the largest model of FB-OCC with InternImage-XL\cite{InternImage} wins the championship of CVPR2023 occupancy prediction challenge.
}

\textbf{Occupancy learning with dense voxel features.} MonoScene\cite{MonoScene} is the first outdoor 3D voxel reconstruction framework based on monocular camera. It uses dense voxel labels from SSC task for evaluation metrics. It includes a module for 2D feature line of sight projection(FLoSP) to bridge 2D and 3D U-Net, as well as a 3D context relation prior (CRP) layer for enhancing learning of contextual information. OccDepth\cite{OccuDepth} is a stereo-based method which lifts stereo features to 3D space via a stereo soft feature assignment module. It uses a stereo depth network as teacher model to distill depth-augmented occupancy perception module as student model. Unlike above methods that require dense semantic voxel labels. 

\textbf{Occupancy learning with sparse-to-dense voxel features.} VoxFormer\cite{voxformer} is a two-stage transformer-based framework, which starts from sparse visible and occupied queries from depth map and then propagate them to dense voxels with self-attention.

\textbf{Occupancy learning with tri-plane features.} TPVFormer\cite{tpvformer} is a the first surround-view 3D reconstruction framework which only uses sparse LiDAR semantic labels as supervision. TPVFormer generalizes BEV to Tri-Perspective View (TPV), which means feature expression of 3D space through three slices perpendicular to the $x,y,z$ axis. It queries 3D points to decode occupancy with arbitrary resolution. For supervision method, TPVFormer\cite{tpvformer} replaces dense voxel grid labels with sparse LiDAR segmentation labels for supervision of dense semantic occupancy from surround-view cameras. Compared to voxel labels with fixed resolution, point cloud labels are more easily accessible, and they can serve as supervision for voxel grids with arbitrary perception range and resolution. 

\new{
\textbf{Extending semantic segmentation to panoptic segmentation for occupancy.} Semantic occupancy prediction performs semantic segmentation on each voxel but lacks instance extraction. Semantic segmentation and instance segmentation are related tasks, and instance information can eliminate inconsistent semantic predictions for one foreground object and mixed predictions for adjacent objects. PanoOcc\cite{PanoOcc} uses detection boxes to split instances from occupancy predictions. Symphonies\cite{Symphonies} and PaSco\cite{PasCo} uses an end-to-end transformer-based panoptic segmentation head to generate instance from multi-scale SSC features. PasCo also proposes voxel uncertainty estimation with model ensemble and demonstrates its robustness on diverse weathers. OccFormer\cite{occformer} proposes a dual-path transformer to encode 3D voxels in different directions and a modified Mask2Former\cite{Mask2Former} for both semantic and panoptic segmentation. 
}
% PanoSSC proposes a novel monocular occupancy network for panoptic 3D scene reconstruction, aiming to predict voxel-level occupancy, semantics and instance id.
\new{
\subsubsection{Fusion-based Occupancy Networks}\label{sec:fusionocc}
OpenOccupancy introduces a multi-modal fusion baseline (M-baseline) for fusion-based occupancy networks. M-baseline consists of SpConv4x lidar encoder, LSS-based occ pooling image encoder, voxel-level fuser and prediction head. M-CONET\cite{OpenOccupancy} extends M-baseline to cascade structure in prediction head and improves accuracy. OccFusion\cite{OccFusion1} is a transformer-based fusion framework which sets LiDAR voxel features as queries and multi-view image features as key and value. OccFusion\cite{OccFusion2} conducts global-local attention voxel-level fusion on multi-scale 3D feature volume. Co-Occ\cite{Co-Occ} adds an additional voxel rendering branch that converts voxel features to a density field and a colour field and calculates loss from depth and pixel color. OccGen\cite{OccGen} proposes a progressive refinement decoder that denoise 3D gaussian noise map to the predicted targets. 
}
% EFFOcc\cite{EFFOcc} is a BEV-based fusion network which gets state-of-the-art performance with help of detection pretraining.
\new{
Multi-agent collabrative fusion is less explored considering the data collection difficulty of vehicle-to-anything(V2X). PointSSC\cite{PointSSC} setup a lidar-based SSC benchmarks built on top of V2X-seq dataset. CoHFF\cite{COHFF} demonstrates significant performance improvement with multi-vehicle hybrid-feature fusion in CARLA simulator.
}

\new{
\subsubsection{Implicit Representation of 3D Occupancy}\label{sec:implicitocc}
Explicit representation of occupancy is structurally efficient on parallelization, but inflexible. Most networks can only work at a fixed perceptual range and resolution after training. Implicit expressions can provide the flexibility to construct occupancy grid at variable resolutions. Signed distance field (SDF) is an important tool to implicitly represent 3D scene in terms of distance from the surface. SIREN\cite{SIREN} propose a weakly-supervised perception that complete 3D mesh from point clouds. LODE\cite{LODE} proposes a generative adversarial network (GAN) to utilize occupancy label to supervised SDF. For implicit learning of vision models without the input of LiDAR, SurroundSDF\cite{SurroundSDF}  proposes the Sandwich Eiikonal formulation for SDF modelling, which provides correct and dense constraints on both sides of the surface. 
}

\new{
\textbf{Analysis: A trade-off between computational demand and precision for 3D occupancy prediction.} It is a common concern that a larger perception range and finer grid resolution leads to higher precision and higher computational demands, and vice versa. Current driving systems usually regards occupancy prediction as a fallback system, with limited perception range of 40m to 70m and limited resolution of 0.2m to 0.4m. Occ3D-Waymo\cite{Occ3D} provides voxel labels with resolution 0.1m but few networks are working on this resolution. Long-range occupancy prediction is an under-explored area. Implicit representation supported variable-resolution networks and multi-resolution (near-small, far-large voxel grids) networks are feasible paths. Grid size is also relevant to the specific engineering design, for example, a fine occupancy grid is required for assisted driving in urban area, but a truck needs a long-distance occupied grid when driving on highways.
}

\subsection{Deployment-friendly 3D Occupancy Networks}\label{sec:deployocc}
\subsubsection{Industrial Design of 3D Occupancy Network}\label{sec:industrialocc}
\begin{figure}[ht!]
	\centering	\includegraphics[width=0.4\textwidth]{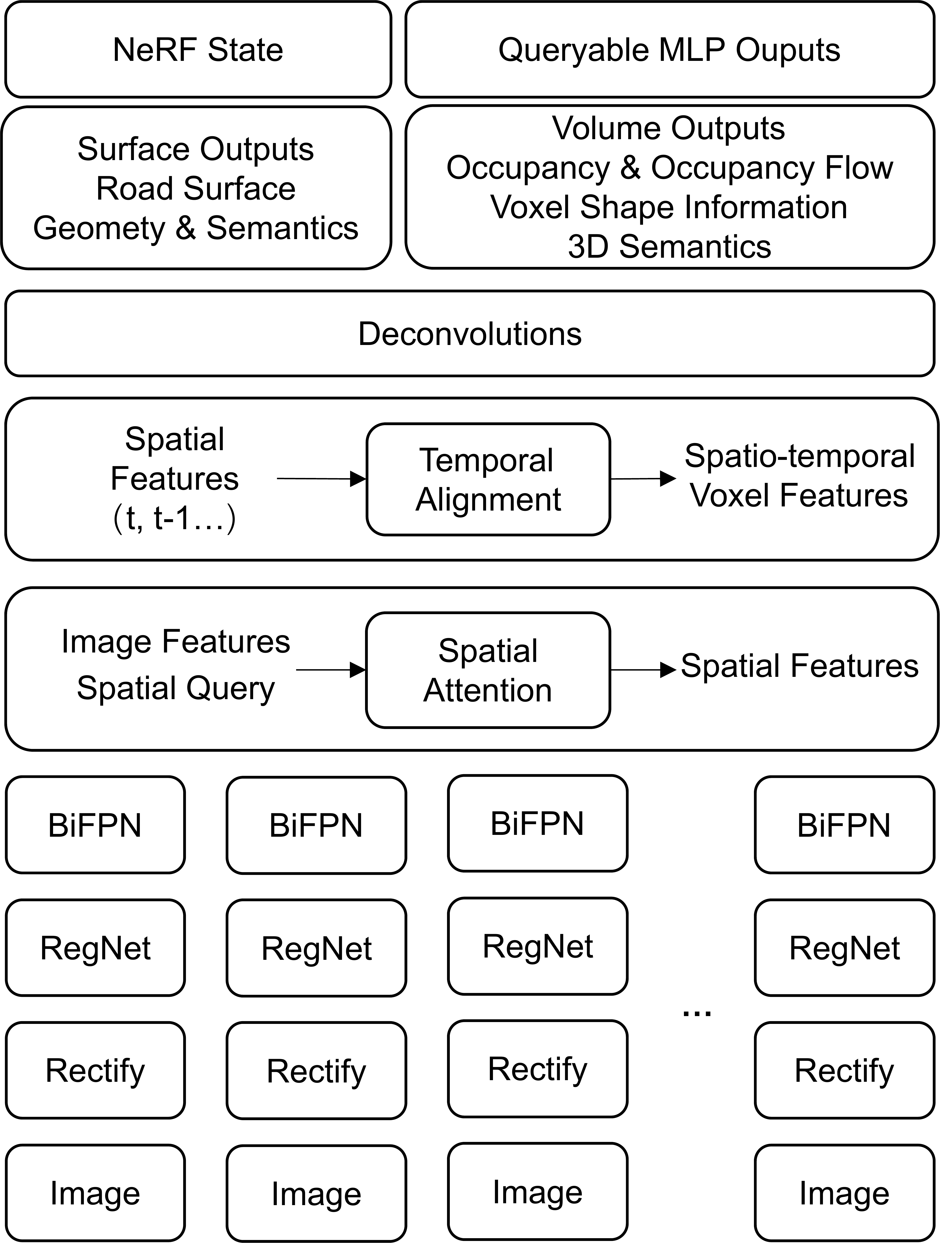}
	\caption{Tesla FSD\cite{tesla_ai_day_2022}: Tesla FSD Beta perception system built on top of Occupancy Network.}
	\label{tesla_fsd_beta}
\end{figure}

Tesla is a pioneer in the investigation of real-time occupancy networks with high performance and low latency at $10ms$ on embedded FSD computers. Tesla first introduces Occupancy Network\cite{tesla_wad22_cvpr} at the CVPR2022 Workshop on Autonomous Driving (WAD)\cite{wad22_cvpr}, followed by the entire pipeline of grid-centric perception system at Tesla AI Day 2022\cite{tesla_ai_day_2022}. The Occupancy Network's model structure is depicted in Fig.\ref{tesla_fsd_beta}. First, the model's backbone uses RegNet\cite{RegNet} and BiFPN\cite{EfficientDet} to obtain features from multiple cameras; then, the model performs an attention-based multi-camera fusion of 2D image features through spatial query with 3D spatial position. The model then executes temporal fusion by aligning and aggregating 3D feature space according to the provided ego pose. After fusing features across temporal horizons and deconvolution modules, the decoder decodes both volume and surface states. The combination of voxel grids and neural implicit representation is also noteworthy. Inspired by NeRF, the model concludes with an implicit queryable MLP decoder that accepts arbitrary coordinate values $x,y,z$ to decode information regarding the position of that space, i.e. occupancy, semantics, and flow. In this way, Occupancy Network is able to achieve arbitrary resolution for 3D occupancy mapping.

\subsubsection{Computation-efficient 3D Occupancy}\label{sec:compute-effi-occ}
\textbf{Efficient View Transformation from PV to BEV}\label{sec:pv2bev-effi}
Vanilla LSS needs a complex voxel cumsum trick to splat probabilistic depth feature on BEV space. Later works\cite{GKT,BEVFusion,BEVDepth} primarily optimizes the computation cost of vanilla LSS\cite{LSS} in designing efficient operators on voxel grids. LSS\cite{LSS} leverages a cumsum trick which sorts frustum features to their unique BEV IDs, which is inefficient in sorting process over BEV grids. BEVFusion\cite{BEVFusion} proposes an efficient, exact without approximation, BEV pooling by precomputation of grid indices, and interval reduction via a specialized GPU kernel that parallelizes over BEV grids. BEVDepth\cite{BEVDepth} proposes efficient voxel pooling which assigns each frustum feature a CUDA thread and corresponds each pixel point to that thread. GKT\cite{GKT} leverages the geometric priors to guide the transformer to focus on discriminative regions, and unfolds kernel features to generate BEV representation. For fast inference, GKT introduces a look-up table indexing for camera’s calibrated parameter-free configuration at runtime. MatrixVT\cite{MatrixVT} observe that the sparsity of feature transporting matrix (FTM) is the root cause of inefficiency. The FTM is orthogonally decomposed into two independent matrices by the Ring \& Ray decomposition method, which encode the distance and direction of the polar coordinate of the auto-vehicle respectively. After Ring \& Ray decomposition, VT can be rearranged into a mathematically equivalent but more efficient formula. MatrixVT reduces the GPU memory and computation of the VT by hundreds of times. 
Fast-BEV\cite{Fast-BEV} is a powerful real-time BEV algorithm based on M2BEV\cite{M2BEV} which proposes two acceleration designs: pre-computing the projection index and projecting
to the same voxel feature. WidthFormer\cite{WidthFormer} address fast view transformation with a single layer of a transformer decoder to compute
BEV representations and a novel Reference Positional Encoding (RefPE). 

% Network details of GKT and BEVFusion are shown in Fig. \ref{gkt_efficient_transformer}, and Fig. \ref{bevfusion_bev_pooling}.

% \begin{figure}[ht!]
% 	\centering
% 	\includegraphics[width=0.45\textwidth]{fig/gkt_efficient_transformer.jpg}
% 	\caption{GKT\cite{GKT}: Efficient kernel for transformer-based PV to BEV transformation.}
% 	\label{gkt_efficient_transformer}
% \end{figure}

% \begin{figure}[ht!]
% 	\centering
% 	\includegraphics[width=0.45\textwidth]{fig/bevfusion_bev_pooling.jpg}
% 	\caption{BEVFusion\cite{BEVFusion}: Illustration of hardware-friendly implementation of BEV pooling introduced in LSS\cite{LSS}.}
% 	\label{bevfusion_bev_pooling}
% \end{figure}

\new{
\textbf{Efficient occupancy networks.}
For the purpose of deploying the occupancy network on-board, researches\cite{FastOcc, SparseOcc_opendrivelab, SparseOcc_noaharklab, FlashOcc} propose different lightweight structure and acceleration methods based on existing occupancy network. From the model perspective, an important improvement is to replace dense Conv3D with lightweight operators\cite{FlashOcc, Fast-BEV}. From the data perspective, more than 80\% of the grid in the vehicle's surroundings is air. PanoOcc predicts a nonempty mask at each layer, performs sparsity on the raster, and then replaces the dense convolution with a sparse convolution. SparseOcc\cite{SparseOcc_noaharklab} (SJTU) removes 80\% of rasters with zero values after LSS view transformation, and SparseOcc\cite{SparseOcc_opendrivelab} (OpenDriveLab) is used to execute Transformer on sparse voxels.
}

\new{
\textbf{Analysis: comparison between feature representation on volume voxel grids and BEV grids.} Voxel features consume much more memory and computing loads than BEV features. For real-time application of 3D voxel grid-based tasks, a compromised approach is to treat vertical dimension as latent features of BEV pillar grids. This 'height as channel' approaches sometimes achieves even better then dense voxel approach. One possible reason is that the empty voxel occupies covers most of the surroundings and voxel features are impaired by the huge imbalance.
}
\section{Temporal Grid-Centric Perception}\label{sec:temporal}

Since autonomous driving scenarios are temporally consecutive and sequential information are natural enhancement of observations of the real world, using spatio-temporal features to decode motion cues and predict future states are important issues for grid-centric perception. 

\subsection{Temporal Fusion for Sequential BEV Features}\label{sec:temporal_bev}
LiDAR pointclouds can easily conduct temporal fusion by transforming all pointclouds to the current frame, but temporal fusion need more tricks for vision models. For LiDAR models, researchers find that using lidar sequences more than 1.5s decreases the performance. To this end, D-Align\cite{D-Align} devises a novel dual query co-attention network and a temporal context-guided deformable attention to achieve inter-frame feature alignment. STEMD\cite{STEMD} introduces two-stage temporal fusion, BEV feature fusion with ConvGRUs and object query fusion with attention. 

For vision models, a common practice is to warp BEV features to the current frame according to relative multi-frame ego poses and undergo temporal blocks. Early arts\cite{STSU,HDMapNet,FIERY} uses simple convolutional blockes for temporal aggregation. BEVDet4D\cite{BEVDet4D} concatenates the wrapped spaces together and undergoes a customized ResNet layer. BEVFormer\cite{BEVFormer} uses deformable self-attention to fuse wrapped BEV spaces. UniFormer\cite{UniFormer} argues wrapped-based methods are inefficient serial methods that do not support long-range fusion, and loses valuable information at the edge of perception range. To this end, UniFormer proposes attention in virtual views between current BEV and cached past BEV which can fuse larger perception range and better model long-range fusion. \new{In general, using more past frames significantly increases training and inference time. To this end, SOLOFusion\cite{SoloFusion} pioneers a streaming training method that only updates the past state with the current frame at a time, so that the full sequence is continuously used for the perception of the current frame.}

\subsection{BEV Motion and Flow Prediction}\label{sec:short-term-motion}

The advantage of direct speed inference on BEV grid is that the speed estimation is at an earlier stage and more stable than that of the object detection and tracking pipeline, and is not affected by recognition failure in detection and the association failure in tracking. The main challenge of motion estimation is that, there is no explicit correspondence relation exists for grids.

\textbf{Ground-truth Labels Generation.}
Common practices to generate the labels of grid flow (scene flow) comes from post-processing of adjacent frames of 3D bounding boxes with unique instance ids. All tracked bounding boxes are transformed to the current frames, assigned unique instance token and mapped to BEV grids as ground-truth labels. 

\textbf{LiDAR models.}
Point clouds lies naturally in 3D space and can be aggregated on data level. The aggregation needs accurate positioning, which may be collected from high-precision GNSS equipments or point cloud registration method(e.g. ICP\cite{ICP}, NDT\cite{NDT}), to convert the point cloud coordinates to the current ego vehicle coordinate system. The feature extraction backbone, with multi-frame point clouds as input, is able to simultaneously extract information in both spatial and temporal dimension to reduce computation load. A compact design is to voxelize pointclouds, treats pointclouds as pseudo BEV maps and vertical information as features on each BEV grids\cite{MotionNet,PillarMotion,BESTI}. MotionNet proposes a lightweight and efficient spatio-temporal pyramid network (STPN) to extract spatio-temporal features. BE-STI proposes TeSE and SeTE to perform bidirectional enhancement of features. TeSE is for spatial understanding of each individual frame. SeTE is for high-quality motion cues by spatial discriminative features. Decoders for motion inputs BEV feature from spatial-temporal backbones. The heads consists of 1-2 stacks of ConvBlock. The predicted heads in MotionNet\cite{MotionNet} include cell classification head for category estimation, motion head for velocity estimation and state estimation head for classifying dynamic or static grids. BE-STI\cite{BESTI} features class-agnostic motion prediction head, which further exploits semantics to more accurate motion prediction. 

For loss design, the spatial regression loss regresses the motion displacement in a L1 or MSE norm manner. Cross entropy loss is used for classification. As consistency is inherently guaranteed by sequential data, MotionNet\cite{MotionNet} proposes a spatial consistency loss for cells belonging to the same object, and foreground temporal consistency loss for temporal constraint on motions between two consecutive frames. As a self-supervised framework, PillarMotion\cite{PillarMotion} proposes a self-supervised structural consistency loss to approximate pillar motion field and cross-sensory loss as an auxiliary regularization to complement the structural consistency given sparse LiDAR inputs.

\new{
The mutual supervision of LiDARs and cameras is effective for learning geometry and motion. PillarMotion\cite{PillarMotion} computes pillar motion in LiDAR branch, and optical flow compensated by ego pose. The optical flow leads to probabilistic motion masking. The optical flow and pillar flow undergo the cross-sensor regulation for a better structural consistency. Fine-tuning of PillarMotion also improves the semantics and motion of BEV grids.
}

\textbf{Vision models.}
Vision models start from image backbone and view transformation module for all frames. Then the temporal module fuse past multi-frame BEV features as described in section \ref{sec:temporal_bev}. Occupancy flow prediction needs state representation of future BEV features. Main components of prediction module are variants of recurrent neural networks (RNN). FIERY\cite{FIERY} proposes Spatial Gate Recurrent Unit (SpatialGRU) for propogating current BEV state to the near future. ST-P3\cite{ST-P3} proposes Dual Pathway Probabilistic Future Modelling (Dual-GRU) which inputs two different distribution of current BEV states for stronger features of prediction. BEVerse\cite{BEVerse} features  iterative flow for efficient future prediction, which inputs last-frame BEV feature to current-frame prediction. StretchBEV adopts a variantional autoencoder from neural ordinary differential equations (Neural-ODE) to learn temporal dynamic through a generative approach. Despite temporal modules design, video transformers also acts as good temporal predictors. TBP-Former\cite{TBPFormer} proposes a spatio-temporal pyramid prediction model based on Swin Transformer, which comprehensively extract multi-scale BEV features and predicted the future BEV state. FipTR\cite{FipTr} proposes a flow-aware BEV predictor for future BEV feature prediction composed of a flow-aware deform-able attention that takes backward flow guiding the offset sampling. 

The prediction head for vision consists of a lightweight BEV encoder (e.g. ResNet18\cite{ResNet}) and four BEV decoders. The five independent decoders output centerness, BEV segmentation, offset to the centers,and future flow vectors, respectively. The post-processing unit associates offsets with centers to form an instance from segmentation and outputs an instance flow from multi-frame instances. The spatial regression loss regresses the centers, offsets and future flows in a L1 or MSE norm manner. Cross entropy loss is used for classification. The probabilistic loss regresses the Kullback-Leibler divergence between BEV features. 

\subsection{Expanding Occupancy Flow to Long-term}\label{sec:occu_flow}
% TODO: Rewrite this part to one's own word
Given ground-truth histories as input, long-term occupancy flow serves as a useful supplementary to trajectory function. Occupancy flow tasks trace occupancy from far-future grids to current time locations using sequential flow vectors. DRF\cite{DRF} uses auto-regressive sequential networks to predict occupancy residuals. ChauffeurNet\cite{ChauffeurNet} supplements safer trajectory planning with a multi-tasking learning of occupancy. Rules of the Road\cite{RulesOfTheRoad} proposes a dynamic framework to decode trajectories from occupancy flow. MP3\cite{MP3} predict motion vectors and their corresponding possibility of each grid. The top three participants of the Waymo occupancy flow challenge are HOPE\cite{HOPE}, VectorFlow\cite{VectorFlow}, and STrajNet\cite{STrajNet}. HOPE\cite{HOPE} is a novel hierarchical spatio-temporal network with a multi-scale aggregator enriched with latent variables. VectorFlow\cite{VectorFlow} benefits from combining vectorized and rasterized representation. STrajNet\cite{STrajNet} features interaction-aware transformer between trajectory features and rasterized features. 

\new{
\subsection{4D Occupancy Forecasting}\label{sec:4docc}
4D occupancy forecasting is formulated to predict the future occupancy state, so this task is also named as predictive world model. Most methods factor out the influence of future ego trajectory by transforming future labels to the current coordinate, except that OccWorld\cite{OccWorld} jointly predicts future occupancy at target coordinate and future ego vehicle trajectory.
}

\new{
One simple approach to start the task of 4D occupancy forecasting is an dimension extension of BEV occupancy flow. Occ4cast\cite{Occ4cast} proposes the first 4DOcc benchmark OCFBench. Occ4cast uses binary voxelized point cloud as input and validate the performance of three baseline temporal baselines: module in 4d-occ-forcasting\cite{4d-occ-forecasting}, LSTM and Conv3D module. Cam4DOcc uses multi-view images as input and validate the performance of four baselines: static model, point cloud forecasting, BEV instance prediction with 3D lifting and end-to-end forecasting with Conv2D temporal predictor. ViDAR\cite{ViDAR} uses a prediction transformer approach to get future BEV features for decoding.  
}

\new{
Temporal networks are effective in short-term prediction, but their predictions are not multi-modal, i.e. it only predicts one possible future prediction). The rise of generative artificial intelligence (GenAI) brings new ideas to multi-modal 4D forecasting. In the context of GenAI, 4D occupancy forecasting is formulated as video prediction task where every pillar is regarded as one pixel. COPILOT4D\cite{COPILOT4D} is a two-stage generative framework. The first stage uses similar BEV encoder, quantizer and decoder with UltraLiDAR\cite{UltraLiDAR}. The second stage applies diffusion strategies\cite{latentdiffusionmodel} on maskgit\cite{maskgit} transformer. COPILOT4D uses three masking strategies to learn prediction tasks. The results show 65\% reduction in 1s prediction and produces diverse future in 3s prediction. As a similar approach, OccWorld\cite{OccWorld} replaces time-consuming diffusion process to autoregressive transformer to produce future prediction in real time. 
}

\section{Label-Efficient Learning for Grid-Centric Perception}\label{sec:effilearn}

Since occupancy label annotation is expensive, label-efficient techniques are urgently needed for learning occupancy networks with easy-to-get weak labels or no labels. On the other hand, practices demonstrate that when one model is pre-trained in an occupancy context, the pre-training strategy can improve the model on all perception tasks.

\new{
\subsection{Occupancy as Pre-training for Downstream Tasks}\label{sec:occ_as_pretrain}
With the great success of large-scale pre-training in the field of natural language processing (NLP), self-supervised visual learning has received wider attention. In 2D domain, self-supervised models based on contrastive learning\cite{MocoV3}, based on masked image modeling\cite{MAE,SimMIM} are developing rapidly and is able to even surpass fully-supervised conterparts. In 3D domain, self-supervised pre-training have been conducted on LiDAR pointcloud\cite{PointContrast,DepthConstrast,PointMAE,ProposalContrast}. The core issue for self-supervised tasks is to design a predefined task for stronger feature representation. The predefined task may originate from temporal consistency, discriminative constractive learning and generative masked learning. 
}

\new{
Researchers find that compared to SSL on raw sensor data, using occupancy prediction as a pretext pre-training task is a more robust and generalizable pre-training technique for various downstream tasks. SPOT demonstrates occupancy pre-training is an general and scalable paradigm that is transforable between different dataset, as occupancy is not limited to the specific LiDAR and occupancy can be formulated as the same between different datasets. Therefore, 2D or 3D grids serve as satisfactory intermediate for self-supervised learning 3D geometry and motion. Voxel-MAE\cite{VoxelMAE1,VoxelMAE2} defines a voxel-based task which masks 90\% of nonempty voxels and aims at completing them. This pre-training has boosted performance for downstream 3D object detection. Similarly, BEV-MAE\cite{BEV-MAE} proposes to mask BEV grids and recover them as a predefined task. MAELi\cite{MAELi} distinguishes between free and occluded space and leverages a novel masking strategy to fit the inherent spherical projection of LiDARs. MAELi shows a significant improvement on performance of downstream detection tasks compared to other MIM-based pre-training. ALSO\cite{ALSO} sets a novel pre-defined task which predicts the 3D occupancy of query points which are sampled along each ray from the origin to the reflected point. For each ray two points close to the reflected point, one outside as free and one inside the surface as occupied, are sampled as query points. This pre-defined task is able to complete the obstacles' surfaces and shows improvement in both 3D detection and LiDAR segmentation tasks. 
}

\new{
Despite LiDAR networks learning from occupancy pre-training, vision networks benefits more from 3D and 4D occupancy learning. UniScene\cite{UniScene} and UniWorld\cite{UniWorld} uses voxelized pointclouds at the current frame and future frames as occupancy supervision and improves downstream BEV detection task. DriveWorld\cite{DriveWorld} learns future occupancy labels with state-space-models and improves all driving tasks such as detection, tracking, planning. ViDAR\cite{ViDAR} is a transformer-based predictor that learns future pointclouds with latent BEV rendering and significantly improves all downstream driving tasks. 
}

\new{
\subsection{Self-supervised 3D Occupancy Prediction}\label{sec:ssl_occ3d}
Fully-supervised 3D occupancy networks are highly dependent on training labels built upon expensive LiDAR segmentation labels, which has very limited scalability to extend to autonomous vehicles without LiDARs. To this end, weakly supervised and self-supervised occupancy networks are proposed to get rid of supervision from other sensors.
}

\new{
Depth estimation is an important prerequisite task for occupancy prediction. Self-supervised depth estimation has a long tradition in 3D vision field. For monocular case, MonoDepth2\cite{MonoDepth2} jointly predicts ego pose and depth map from monocular videos in a novel view synthesis manner. For multi-view case, SurroundDepth\cite{SurroundDepth} uses cross-view transformer(CVT) to capture clues between different cameras and uses pseudo depth from structure from motion operators. NeRF\cite{NeRF} is also a promising approach for geometric self-supervision of depth. Behind the scenes\cite{BehindTheScenes} and SceneRF\cite{SceneRF} proposes novel depth synthesis by refining a MLP radiance density field which can infer depth of source frame image with other frames in one sequence. Depth network trained with neural rendering demonstrates less relative and absolute depth errors.
}

\new{
Reconstruction with images is a extensively problem in the field of 3D geometry. Injecting geometric knowledge into real-time perception networks is an emerging challenge. Neural rendering serves as a weak signal from 2D information to supervise 3D geometry. The basic idea is to construct a density field and a semantic field from voxel features with MLPs, sample attributes along the rays and integrate the attributes to the image plane and calculate losses with image plane information. UniOcc\cite{UniOcc} and RenderOcc\cite{RenderOcc} uses image RGB pixels, depth from LiDAR and segmentation from LiDAR as weak supervision which improves occupancy performance. S4C\cite{S4C}, SelfOcc\cite{SelfOcc} and OccNeRF\cite{OccNeRF} uses video sequences only for geometric reconstruction with NeRF. Moreover, OccNeRF uses grounding-dino with segment-anything model(SAM)\cite{SegmentAnything} to generate high-quality image segmentation. As multi-view stereo has a strong limitation of static scene, most NeRF-based SSL uses static categories from adjacent frames and dynamic categories only at the current frame. This approach greatly limits the geometric reconstruction for dynamic objects. To make better use of object trajectories, OccFlowNet\cite{OccFlowNet} pre-computes scene flow from ground-truth object trajectories and then move dynamic voxel indices with flow offset and then use RenderOcc\cite{RenderOcc} as weakly-supervised occupancy learning. 
}

\new{
\subsection{Self-supervised 4D Occupancy Forecasting}\label{sec:ssl_occ4d}
4D forecasting labels are expensive, and in most cases impossible to collect as sensor data only reflects the surface of the world. To this end, 4d-occ-forecasting uses future point clouds as a proxy for future occupancy and thus allows the evalaution on any dataset with LiDAR pointcloud sequences. The idea is simple that given the future sensor origin, the predicted future point clouds is the ground-truth ray projected to the nearest future predicted as occupied area. They proposes a differential voxel rendering layer to calculate loss by comparing ground-truth ray length and predicted ray length in the same direction. ViDAR\cite{ViDAR} further develops the binary voxel ray rendering to BEV latent ray rendering, which is to integrate voxel values along the ray as the length of the ray. Compared to 4d-occ-forecasting, ViDAR's approach does not explicitly construct a binary 4D occupancy grid, but it contributes more as powerful BEV pre-training strategies to enhance downstream tasks performance.   
}

\new{
\subsection{Open-vocabulary 3D Occupancy Prediction}\label{sec:ov_occ}
Most occupancy networks predicts semantic categories in a close-set plus one general object as others, but the variety of obstacles that appear on the road is endless. Open-vocabulary occupancy netowrks aims to detect any obstacles given text prompts as input. A common approach is to distill voxel feature to the vision-language teacher model, e.g. CLIP and its varients. OVO\cite{OVO} is the first attempt of open-vocabulary SSC with help of high-quality voxel selection. POP-3D\cite{POP-3D} is built on TPVFormer\cite{tpvformer} and conducts tri-plane self-supervised training from a MaskCLIP\cite{MaskCLIP} teacher model.
}

\section{Grid-Centric Perception in Driving Systems}\label{sec:driveapplication}

\new{
As a high-level representation of the spatial environment, an occupancy grid implicitly includes rich perceptual information as a basis of planning in autonomous driving, such as collision-free drivable areas, road structure information, spatial risk distribution, and so on. 
}

\new{
Specifically, grid-centric perception results can serve as trajectory evaluators by quantifying the cost or performing safety checks on the planning results. Furthermore, by modeling the objective functions and constraints based on the grid, it can also be used as for trajectory optimization. Additionally, using rasterized environmental representations as input for driving policy networks is also a common planning paradigm, which allows for a unified description of the environment. 
}

\new{
Due to the dense representation and rich spatial semantic description capabilities of the grid, it showcases generality, spatial details, interpretability and other advantages in the downstream tasks of autonomous driving.
}
% Gird-centric perception provides downstream modules of autonomous driving with rich perceptual information and intuitive collision risk. This section introduces how grid-centric perception benefits downstream planning tasks.

\new{
\subsection{Grid-Centric Trajectory Evaluation}\label{sec:grid_traj_eval}
A grid-based environmental representation encodes safety-related information of discrete spatial locations, such as agent occupancy and road areas. Consequently, it is inherently suitable for serving as an evaluator for waypoints, used for conducting collision checks on sampled trajectories to filter out infeasible ones, or for scoring trajectory candidates to select the optimal solution. 
Conventional rule-based planning algorithms, like sampling-based\cite{chen2022rrt, Lee2019collision} or searching-based\cite{montemerlo2008junior} methods, usually utilize occupancy grids to check the trajectory collision risks. 
Recently, learning-based approaches has emerged to construct a grid-wise environment representation as an evaluator to score a bunch of trajectory candidates. NMP\cite{NMP} proposes an interpretable end-to-end planner where the backbone network outputs a cost volume to represent spatio-temporal occupancy and serve as a safety indicator to select from trajectory sample. Similar planning frameworks are adopted by P3\cite{p3}, MP3\cite{MP3}, ST-P3\cite{ST-P3}, OccNet\cite{OccNet}, which all construct a temporal series of semantic occupancy grids indicating dynamic objects and static maps, and utilize them to design cost functions evaluating sampled trajectory candidates. Instead of predicting an explicit dense 3D-grid map, QuAD\cite{QuAD} leverages an implicit occupancy model that takes BEV latent representation as input to query interested waypoints' cost, thus releasing great computational burden. UAP-BEV\cite{UAP-BEV} learns additionally the uncertainty of BEV prediction and samples perturbation to enhance driving safety. 
}

\new{
Furthermore, another common application of occupancy in AD planning involves the construction of a safety-aware loss during training. For instance, PLUTO\cite{pluto} penalizes those trajectories beyond grid-alike drivable areas with an auxiliary loss, while NMP\cite{NMP} incorporates a max-margin loss to encourage the expert trajectory to have the minimum cost. Sine the loss formulation is also a criterion on the planning results, such methods are also categorized under trajectory evaluation.
}

\new{
\subsection{Grid-centric Trajectory Optimization}\label{sec:grid_traj_optim}
Different from the method of evaluating sampled trajectories, trajectory optimization refers to constructing a numerical optimization problem based on grid-centric environmental representation to solve for the optimal trajectory. 
For this non-convex and nonlinear problem, most algorithms require an initial solution as a starting point for optimization. 
Consequently, many traditional motion planning algorithms optimize the trajectory based on the initial path provided by sampling algorithms like RRT, in order to ensure the kinematic feasibility of the trajectory and provide interpretable safety guarantees. 
It generally requires a differentiable occupancy grid\cite{cho2023model}, which is often referred to as a risk potential field\cite{lu2020adaptive, raksincharoensak2016motion} or a heatmap as well.
}

\new{
For the same reasons, a trajectory optimizer is often employed as a post-processing component of the neural planner for trajectory refinement in many learning-based planning frameworks.
Based on occupancy predictions, UniAD\cite{UniAD} and Hoplan\cite{hoplan} both leverage Gaussian distribution to formulate an objective function concerning collision risk, which is used in optimization to enhance safety of the initial trajectories generated by the network.
Gameformer Planner\cite{gameformerplanner} defines an occupancy grid which represents the temporal occupancy only in the longitudinal dimension, and ensures the vehicle's position not exceeding the safe areas in post-optimization. OPGP\cite{opgp} derives a potential collision cost from occupancy predictions with a penalty on longitudinal distance below the safety threshold.
}

\new{
\subsection{State Representation for NN Policies}\label{sec:grid_state_repre}
Grid-centric perception results consistently describe the spatial information of the environment and are often directly used as inputs for neural network (NN) planners. They are typically organized as multi-channel raster images, with each channel having different physical meanings or representing different historical steps.
For example, \cite{Isele2018} wraps observation as a three-channel grid image representing respectively occupancy, velocities and headings, followed by \cite{saxena2020driving} with another displacement channel.
\cite{chen2023attention} defines a normalized multi-channel grid state where different kinematic features scale in the given ranges.
And \cite{rezaee2021motion} takes the occupancy of two consecutive frames as state input.
Besides, TOFG\cite{wen2023tofg} proposes to constructs a temporal occupancy flow graph for a GAT-based model, where the connections are explicitly modelled between girds as well as between steps.
ChauffeurNet\cite{ChauffeurNet} renders all the observation into rasterized BEV images as the model input, including the roadmaps, routes, dynamic boxes, etc.
Instead of a high-resolution grid, You et al.\cite{you2019advanced} focus on the occupancy of nine surrounding grid cells with the coarse-grained size of a vehicle for state representation.
}

\new{
\subsection{Grid-Centric Representation in Generative World Model}\label{sec:simworld}
Latest world models\cite{DrivingDiffusion, DriveDreamer, DriveDreamer2, GAIA-1, GenAD-OpenDriveLab} in autonomous driving field can simulate the future multi-view images with a generative framework. They typically use high-definition vector map and 3D bounding boxes as control inputs in a diffusion framework. However, these compact elements controls limited variety of scenes elements. Occupancy as intermediate representation or final outputs can introduce more variety to simulation. XCube\cite{XCube} proposes a novel generative method featuring sparse voxel hierarchy to generate high-resolution meshes from single-frame point clouds on waymo dataset. WoVoGen\cite{WoVoGen} uses 3D occupancy labels as world volumes and jointly generate future occupancy and future multi-view images. UrbanDiffusion\cite{UrbanDiffusion} proposes a novel diffusion model which uses BEV layout as input and generate unbounded urban scenes at the 3d occupancy level to preserve the scene geometry and semantic information. SemCity\cite{SemCity} proposes a tri-plane diffusion network that can both inpaint and outpaint to large-scale city semantic scene completion.
}

\new{
These world models are often integrated with model-based planning methods, using data augmentation to learn driving policies in imagination. For instance, MILE\cite{MILE} proposed a model-based IL framework that generates future states and actions iteratively which can be decoded into BEV occupancy, while Think2Drive\cite{think2drive} utilizes the model to synthetize experience data for RL training.
}

\new{
\textbf{Analysis: the advantages of grid-centric representation in planning over object-centric results.} Compared to object-centric pipelines, grid-centric perception does not require explicit object uncertainty modeling and naturally represents the uncertainty of the full space; Compared with the direct regression of vectorized trajectory in prediction or planning, it can inherently deal with multi-modal problems. As a state representation, it eliminates the problem of object lists such as arrangement invariance\cite{leurent2018survey}; In addition to the aforementioned advantages, planning on grid-centric perception is also highly interpretable.
}

\section{Discussion}\label{Discussion}
In this section, we present an in-depth summary of the current trend of grid-centric perception and provide a few future outlooks for the directions of further development. 

\subsection{Observation and Limitations of Current Trend}

Compared with object-level perception pipelines, grid-centric methods have few geometric assumptions of how objects are shaped and greater flexibility to describe objects under any occlusion. Grid-centric methods have become an integral component of vehicle perception systems. Three perspectives summarize the current trends.

\new{
\textbf{Feature representation on 2D/3D/4D occupancy grids.} Compared with the conventional probabilistic-based OGM, deep learning has substantially improved the ability to describe the semantics and motion of grids. The ability to represent features is heavily dependent on the network structure. The representation of features from LiDAR, vision, and radar raw data to BEV and 3D occupancy grids has been investigated extensively in the past two years. Temporal modules are more investigated in the context of history information fusion than future frame predictions. 
}

\new{
\textbf{Efficient learning.} In mainstream supervised occupancy learning, the current label production cost is high, which limits the upper limit of occupancy model. As grid labels are generally expensive, label-efficient learning, such as semi-, weakly-, or self-supervised learning, which is still in its infancy in 3D domains, is anticipated to accelerate the development of future solutions to handle open-world traffic scenarios. There is a growing trend of occupancy as pre-training task to provide an foundation model for performance improvement of downstream perception and planning tasks.
}

\textbf{Applications in driving systems.} We observe that grid-centric perception applications are playing an increasingly crucial role in autonomous driving systems as a whole. Grid-centric perception has a great capacity for obstacle representation and intuitive indication of collision risk. There is a long history of grid-dependent planning for downstream tasks. An emerging trend is that end-to-end planning methods exhibit strong potential for conveying grid features to environment cognition modules and constructing more accurate safe fields.

\new{
\textbf{Limitation of current approaches.} 
In addition to the rapid progress in this area, we also summarize the limitations of current methods. 
}

\new{
In 3D occupancy prediction, scalable and affordable occupancy label generation pipelines are needed. Several techniques may helps, which are neural rendering\cite{Emernerf}, diffusion\cite{WoVoGen} or virtual data from simulation\cite{LightWheelOcc}. For occupancy built upon LiDAR or camera data, the model efficiency needs to be improved with lightweight design to meet the real-time deployment requirements. 3D occupancy networks  with other sensors are rarely explored.
}

\new{
In temporal prediction and 4D forecasting, most occupancy prediction methods focus on  motion estimation at current frame and near future, cause occupancy forecasting primarily performs prediction according to inertia. Predicting the long-term (to 8s) future should consider more social interaction factors, which is the advantages of trajectory prediction. The combination of occupancy and trajectory prediction which balances between inertia of all objects and interaction between known agents remains under-explored.
}

\new{
In label-efficient learning, most self-supervised learning techniques still uses LiDAR points as supervision signals. However, there are huge number of vehicles equipped with multi-view cameras only. Using LiDAR as pre-training targets limits the scalability of both data and model. On the other hand, current occupancy networks which only rely on video sequences have weak performance, which is around $10mIoU$ on Occ3D-nuScenes benchmark, only around 25\% performance compared to fully-supervised vision models. Using large-scale video data without LiDAR to train occupancy networks needs to be explored.
}

\new{
In planning with grid-centric perception, the design of adapters from occupying grids to the latest decision-making techniques is still difficult.  This is mainly caused by the fact that the dimension of occupancy grids is too high. Except for post-processing modules to perform collision avoidance, it remains difficult for planning modules to make full use of rich perceptual information in an occupancy grid. 
}

\subsection{Future Outlooks}

\new{
Grid-centric perception has the potential to create interdisciplinary research opportunities with 3D vision, geometry and mobile robotics. For example, a unified 3D foundation model on occupancy is expected to generalize well on robotics for different purposes. 
}

\new{
\subsubsection{Related Emerging Technologies}
Occupancy task is highly related to 3D vision and 3D geometry. It also has the potential to combine generative ai techniques for prediction and simulation tasks. This section provides a brief introduction of emerging technologies possibly related technologies.
}

\new{
\textbf{Gaussian splatting.} Following the success of NeRF, 3D Gaussian splatting (3DGS)\cite{3dgs} is a novel volumetric rendering technique which represents a 3D scene as a collection of gaussians. The splatting process involves placing these Gaussians at appropriate locations and intensities to novel rendering views to represent the geometry and appearance of the scene. 3DGS renders at a signifantly faster speed and shows better novel view synthesis quality. SuGaR\cite{SuGaR} first proposes surface-aligned 3D-GS to get mesh representation of the scene.
}

\new{
\textbf{Video generation with generative models.} Video generation is an emerging technique for automatically creating video content from image or text. Video generation is highly connected to the concept world model. These models are typically based on deep learning frameworks and utilize techniques such as generative adversarial networks (GANs), variational autoencoders (VAEs),  recurrent neural networks (RNNs), and latent diffusion models to generate sequences of video frames.  Inspired by SORA\cite{SORA} from OpenAI which uses video-transformer based diffusion to generate 60 second geometrically-consistent long videos, OccSora\cite{OccSora} is a 4D world model with trajectory prompts that can produce future occupancy videos to 16 second. Remarkably, combination of video generation and neural rendering will enable the geometrically-consistent and photo-realistic generation of 3D scenes. 
}

\subsubsection{Towards Efficient Representation of 3D Occupancy}

In real-world driving scenarios, nearby surroundings usually have greater risk potentials than distant ones, necessitating increased vigilance as obstacles approach. In grid-centric perception, a natural demand is that nearby grids need higher resolution than distant grids. To step further, grids feature representations are redundant on unnecessarily concerned areas (e.g. distant or occluded regions). Vision transformers and implicit representation are promising techniques for on-demand variable granularity. Another concern is to define the safe granularity for downstream tasks, as it is strongly tied to perception requirement analysis.

\new{
\subsubsection{Towards Scalable Occupancy Pre-training with No Labels}
In comparison to object-centric perception, grid-centric perception requires more rigorous labeling. LiDAR point clouds are constrained by a minimum angular resolution and a maximum height, meaning that distant obstacles may not be accurately marked. Label-efficient learning for 3D vision on grids is urgently needed. Current self-supervised learning is only tested on small datasets and performs very limited precesions. Scaling up of the occupancy pretraining is a promising solution for 3D vision foundation models.
}

\new{
\subsubsection{Towards 4D Occupancy World Model}
4D occupancy is an emerging topic with limited number of practices. Existing approaches, temporal models have difficulty in handling long-term sequences, while generative models have difficulty in real-time deployment. 4D Occupancy world models have great potential of modelling the final output of perception with anything inside. Besides, world models are favourably trained in a self-supervised manner for better scaling up. 
}

\new{
\subsubsection{Towards Open-world Semantic Understanding with Occupancy Prior}
Occupancy grids provides a strong scale-aware geometric piror, which may be helpful for open-world semantic understanding. GOOD\cite{good} is an example of using scale-ambiguous depth to better detect unknown objects. Another approach is to distill open-vocabulary language-aware features to voxel features. On the other hand, occupancy grid has its difficult case when impassable obstacles (e.g., lying down people) conform to the ground, accurate open-world semantic understanding is needed to distinguish between passable free-space with bumps or depressions and impassable pavement anomalies. Reconstruction and semantic understanding can benefit a lot from improvement of each other.
}

\new{
\subsubsection{Towards Better Adapters for Planning with Grid-centric Perception}
Due to the difficulty in extracting instance-level feature information from a raster image, existing methods often lack social interaction processing for planning. How to explicitly or implicitly model the interaction patterns between grid regions is worthy of in-depth consideration.
Besides, grid representation is naturally suitable for modeling uncertainty in perception and prediction, but in fact, very few planning algorithms have focused on this advantage. Contingency planning based on uncertainty-aware occupancy grids could be a promising research topic. Moreover, the neural network policies with a high-dimension grid state representation is likely to encounter difficulty with generalization, as well as great computational burden, especially those with a temporal series of girds.
}

\new{
\subsection{Broader Impacts of Grid-centric Perception} We discuss possible ethical and broader impacts of grid-centric perception. Grid-centric perception may be used in a wide range of automation applications, including mobile robots and autonomous vehicles. Grid-centric perception models have better protection for anonymization and privacy than object detection models, as it does not require the description of each individual to be labeled in detail in the annotation file, and exists in the form of an occupied grid.
}
\section{Conclusion}\label{conclu}
This paper provides a comprehensive review and analysis of the well-established and emergent grid-centric perception of autonomous driving. The background begins with the introduction, datasets, benchmarks, and evaluation metrics for grid-centric perception tasks. \new{We step from the BEV representation from multi-modal sensors to 3D grid representation. For advances in temporal learning in grid-centric perception, we review history infomation fusion, BEV motion, and 4D occupancy flow. Afterward, we provide a thorough investigation of label-efficient learning to learn occupancy networks with few labels as universal pretraining. We review how planning methods benefit from grid-centric perception results. Finally, we present a summary of current research trend and limitations of current technologies, emerging technologies, future outlooks and broader impacts of grid-centric perception.} The authors hope that this paper will prospect the future development of grid-centric perception on autonomous vehicles.

\appendix{}

\subsection{Occupancy Grid Mapping Fundamentals}\label{appendix:ogm}

\subsubsection{Sensory Inputs for Autonomous Vehicles} Autonomous vehicles rely heavily on multiple cameras, LiDAR sensors, and RADAR sensors for environmental perception. The camera system may consist of monocular cameras, stereo cameras, or both. It is relatively cheap and provides high-resolution images $I_{cam}\in \mathbb{R}^{H \times W \times 3}$, including texture and color information. However, cameras cannot obtain direct 3D structural information and depth estimation. In addition, image quality is highly dependent on environmental conditions.

LiDAR sensors generate a 3D representation of the scene in the form of a point cloud $I_{LiDAR}\in \mathbb{R}^{N \times 3}$, where $N$ is the number of points in the scene and each point contains the $x,y,z$ coordinates, as well as extra attributes such as reflective intensity. Due to depth perception, a broader field of view, and a greater detection range, LiDAR sensors are utilized more frequently in autonomous driving and are less susceptible to environmental conditions. Unfortunately, the applications are mainly limited by cost.

RADAR sensors are one of the most significant sensors in autonomous driving due to their low cost, long detection range, and ability to detect moving targets in adverse environments. RADAR sensors return points containing the relative locations and velocities of the targets. However, RADAR data is sparser and more sensitive to noise. Hence, autonomous vehicles frequently combine RADAR data with other sensory inputs to provide additional geometrical information. 
% It is believed that 4D imaging radar will become a key enabler for low-cost L4-L5 autonomous driving with significant improvements. 4D imaging radar is able to generate dense pointclouds with a higher resolution and estimate the height of objects\cite{TJ4DRadSet,astyx}. Seldom are 4D radar applications applied in grid-centric perception.

\subsubsection{Occupancy Grid Map}\label{appendix:ogm_base}
Originating from the field of robot environment representation\cite{Highresolutionmaps}, \textbf{Occupancy Grid Map (OGM)} are mainly used to deal with the uncertainty of radar measurements and model complex environments robustly and uniformly. Early robots are commonly equipped with single-scan laser or multi-scan LiDAR. Due to unavoidable errors in the radar measurement, OGM is proposed to divide the environment into discrete independent grids and estimate the occupancy probability of each grid through an inverse sensor model\cite{OGMforRobot}\cite{ProbaRobotics}.

The classical OGM describes the grid state $O_k$ at time $k$ as binary opposite events: occupied or free, i.e. $O_k \in\{0, F\}$. When the measurement at time $k+1$ arrives, the OGM updates the state of the grids. For outdoor scenarios, multi-scan LiDAR needs a preprocessing of ground removal by RANSAC\cite{RANSAC} algorithm and mature clustering techniques\cite{Autoware} to acquire the point clouds that only represent the non-ground scene as the input measurement. Afterward, an inverse sensor model assigns discrete binary occupancy probabilities $p_{z_{k+1}}\left(o_{k+1} \mid z_{k+1}\right)$ to each grid based on the measurement $z_{k+1}$ at time $k+1$, this result is known as the measurement grid map.

\begin{figure}[ht!]
	\centering
	\includegraphics[width=0.45\textwidth]{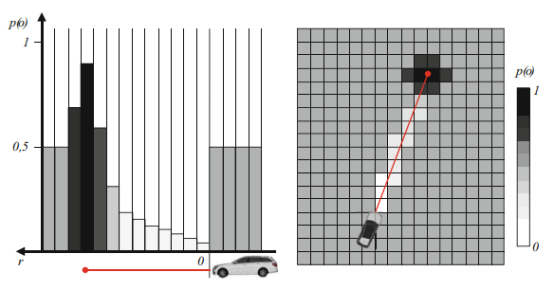}
	\caption{An illustration of an inverse sensor model of a single LiDAR reflection showing the side view (left) and birds-eye view(right)\cite{inverse_sensor_model}}
	\label{inerse_senspr_m}
\end{figure}

A simple example of an inverse sensor is shown in Fig. \ref{inerse_senspr_m}, the occupancy probability of the grid is a function of the distance $r$, assuming that the measurement becomes more uncertain when the distance is greater. Therefore, the inverse sensor model assigns a larger occupancy probability to the grid near the reflection point, and a small occupancy probability to the grid in the area between the robot and the reflection point, which is called free space. This algorithm models each radar beam individually, which does not always match the actual situation. Researchers often design different inverse sensor models according to specific task needs\cite{semantic_evidential_grid}\cite{other_inverse_s_m}, or use machine learning algorithms to derive model parameters\cite{deep_inverse_s_m}.

OGM updates the posterior occupancy probability following the Bayesian filtering mechanism. Given measurements at time $k+1$ from sensor inverse model, the binary state of each grid at time $k+1$ is estimated via a binary Bayesian filter(BBF)\cite{ProbaRobotics}:
\newenvironment{sequation}{\small\begin{equation}}{\end{equation}}
\begin{sequation}
  \begin{aligned}
&p_{o, k+1}\left(o_{k+1}\right)= \\
&\frac{p_{z_{k+1}}\left(o_{k+1} \mid z_{k+1}\right) \cdot p_{o, k}\left(o_k\right)}{p_{z_{k+1}}\left(o_{k+1} \mid z_{k+1}\right) \cdot p_{o, k}\left(o_k\right)+p_{z_{k+1}}\left({f}_{k+1} \mid z_{k+1}\right) \cdot p_{o, k}\left({f}_k\right)}
\end{aligned}
\label{BBF}
\end{sequation}
\begin{footnotesize}
\begin{multline}
p_{o, k+1}\left(o_{k+1}\right)=\\
\frac{p_{z_{k+1}}\left(o_{k+1} \mid z_{k+1}\right) \cdot p_{o, k}\left(o_k\right)}{p_{z_{k+1}}\left(o_{k+1} \mid z_{k+1}\right) \cdot p_{o, k}\left(o_k\right)+p_{z_{k+1}}\left({f}_{k+1} \mid z_{k+1}\right) \cdot p_{o, k}\left({f}_k\right)}
\end{multline}
\end{footnotesize}
where $p(o)$ and $p(f)$ denote the probability of the occupied or free state, respectively.The preconditions for the establishment of Eqn. \ref{BBF} include: the prior probabilities of occupancy state and free state are equal, measurements of grids are independent of each other, and the state does not change with time.

OGM has the following advantages: (a) a unified probabilistic form for environment perception with a solid theoretical foundation; (b) the ability to model complex obstacles of arbitrary shapes; (c) An intuitive and clear estimation of free space (driveable space); (d)  A convenient framework for the fusion of different sensor data. 

However, fixed-resolution OGM has obvious shortcomings which hinder its application in the field of autonomous driving. (a) Reasonable resolution for balancing precision and computation burden. (b) Sparse and noisy laser input leads to a large number of unknown regions\cite{3D_OGM_Bayesian_kernel}. An important premise for Eqn. \ref{BBF} is that the measurement of the grid is completely independent, that is to say, when there is no radar ray reflection point in a grid, it will be regarded as an unknown state. The sparsity of the point cloud means that most of the grids in the environment are in unknown states. (c) Failure of static OGM to cope with dynamic scenes. Moving vehicles leave trails on static OGM.  

\subsubsection{Continuous Occupancy Map}\label{appendix:com}

The Continuous occupancy map is a split-new approach that uses continuous occupancy probability kernels which allows arbitrary resolutions of occupancy. A continuous occupancy map first divides the point cloud into free or occupied parts and updates the parameters in the kernel function according to these points or segments with known states, then using the obtained kernel function to estimate the occupancy state of any nearby location and realize the continuous occupation rate estimation of the whole space. Gaussian process occupancy maps (GPOMs)\cite{GPOMS} use a modified Gaussian process classifier model as a non-parametric Bayesian technique while introducing correlations between points in the map to model the continuous probability representations of occupancy estimation with correlated variance maps in the real-world environment, especially those areas with sparse point clouds. But its time complexity is $\mathcal{O}\left(N^3\right)$ in the number of training points, which brings difficulties for real-time applications. Hilbert maps\cite{HilbertMaps} project the original data into the Hilbert feature space based on the kernel feature approximation and use stochastic gradient descent to train a logistic regression classifier. This algorithm is robust to outliers and the update time is linear with the growth of data points, obtaining a favorable balance between the continuous occupancy estimation effect and real-time performance, but it cannot calculate the uncertainty of the estimation result like GPOM.

\subsubsection{Dynamic Occupancy Map}\label{appendix:dogm}
In traffic scenarios, occupancy maps need to represent the occupancy of dynamic obstacles such as pedestrians and other vehicles and simultaneously estimate their motion state to further predict future trajectories and avoid driving risks.

The most intuitive method is to combine static grid map with multi-target tracking technology, which uses OGM to represent static obstacles and \textbf{detection and tracking of moving objects (DATMO)} to model dynamic obstacles. Early approaches obtain the range of dynamic obstacles by checking conflicts between grid map and the added observations and these conflicting regions would be used as input to the multi-target tracking algorithm and removed from the static grid map accordingly, yielding a robust static occupied raster map\cite{SLAMandMOT,grid_traking2014,Mapbuilding2003,Online_slam&mot,markov_chainDOGM}. But the main issue with adopting two different environmental representations to model a unified scenario is the difficulty of data association. The static grid structure and the polygon list of dynamic obstacles are likely to be inconsistent when correlating. Meanwhile, the shape assumptions of dynamic obstacles limit the estimation accuracy and application range of the system, which is contrary to the advantage of OGM to represent arbitrary obstacles intuitively. Attempts on the data association issue include clustering conflicting grids and associating them directly to multi-object tracking to implement separate tracking of each part\cite{OGM&MOTfusion}, and constructing exclusive local grid maps for each dynamic object\cite{SLAMandMOT}.

Another key concern of the DATMO-based approach for representing dynamic objects is that both the detection and shape models of dynamic obstacles must be well trained\cite{semantic_evidential_grid}, since the false and missed detections generated at this stage will propagate  with the whole system process in a chain, leading that unknown objects and features in practical applications are probably to cause a significant impact on the performance of the algorithm. Therefore, a more reasonable approach is to directly improve the occupancy map itself in order to achieve a uniform and consistent representation of dynamic and static objects.

\textbf{Dynamic Occupancy Grid Map (DOGM)} integrates dynamic targets into the grid map and estimates the occupancy probability of obstacles corresponding to the grid as well as their velocity state, so as to improve the perception and obstacle avoidance capability of robots or autonomous vehicles for complex dynamic environments.

The methods are varied in the publications. For continuous occupancy maps, some have applied the dynamic obstacle assumption to existing static occupancy graph structures. For example, \cite{DGPOM} used an optical flow-based motion map to estimate the  velocity of the grid and improved the GPOM\cite{GPOMS} to accommodate a dynamic environment. In \cite{Dhilbertmaps}, point clouds in LiDAR are clustered and filtered to estimate the velocity of dynamic obstacles, which is used to generate a non-stationary kernel in Hilbert space to build a dynamic Hilbert map\cite{HilbertMaps}.

Bayesian Occupancy Filter (BOF)\cite{BOF,BOFtracking} is entirely grid-based with no object assumptions. BOF estimates the two-dimensional environment using a four-dimensional grid, where two grid dimensions represent the spatial position and two grid dimensions represent the two-dimensional velocity of the obstacle. Thus, BOF estimates the motion of an object and considers it explicitly in its process model. It uses the histogram of each grid to estimate the discrete grid velocity distribution, which means the velocity is equal to an exact integer grid displacement in the prediction process\cite{BOFtracking}. However, this discretization of the velocity leads to artifacts and limits the accuracy. In addition, each grid requires summing over all possible antecedent grids and their velocities, which is computationally expensive, especially when the environment needs to be expressed with high accuracy.

One class of methods that have emerged in recent years uses particles as the basis for maps, which were designed for dynamic environments from the beginning. In particle-based methods, obstacles are considered as a set of point objects and particles with velocity are used to simulate point objects. The theoretical basis of the particle-based approach originates from the Sequential Monte Carlo (SMC) filter\cite{danescu_particle2010,danescu_particle2011}. The entire dynamic grid is represented by a particle swarm, in which each particle represents an individual hypothesis that can move from one grid to another. Particle states are defined by position and velocity and are predicted across the grid based on their motion model and parameters. The article is directly associated with a grid based on its position, thus contributing to the occupancy and velocity distribution of this grid. The occupancy probability is described by the number of particles in the grid.

Random Finite Set (RFS) theory is introduced in \cite{DOGM} to particle-based maps and derived a map construction procedure using Probability Hypothesis Density (PHD) filters and Bernoulli filters. Dempster-Sheffer (DS) evidence theory is then used to update the occupancy state of a dynamic grid to significantly reduce the computational effort.

In cluttered environments with both dynamic and static obstacles, it is necessary to model dynamic or static multi-point objects and non-negligible noise may occur. To reduce the noise, two approaches are usually used. The first approach is to use a hybrid model \cite{dualPHDfilter,evidentialmap&particletrack,evidentialgrid}, which includes a separate static model and a dynamic model to update the states of static and dynamic point objects independently. This hybrid model can be used as a dual PHD filter\cite{dualPHDfilter} or grid-level inference\cite{evidentialmap&particletrack,evidentialgrid}. Another approach is to apply additional information during the update to reduce noise, including object height\cite{digital_elevation_maps}, grayscale intensity\cite{digital_elevation_maps,grayscale_grid}, semantic information\cite{grayscale_grid} and object id\cite{object_grid_1,object_grid_2}.

In conclusion, the existing particle-based DOGM are promising, but they have two limitations. The first limitation is that most solutions use the totality of a particle (a multi-modal probability density) to estimate the state of multiple grids that are considered independent (a grid is described by its own probability density). A particle can migrate from one grid to another, decoupled from its previous source grid density, and reassigned to a new target grid density. Because particles are recombined into new independent grid estimators in each iteration, the identity of the tracked grid is lost. However, if additional heuristics are used, this information can be retained. In other words, particles can usually identify the presence of an object, but not the presence of a "specific" object part or grid. In order to improve estimation and data association, the authors of \cite{object_grid_1,object_grid_2} adopted the idea of connecting particles to objects by extending them with object ids. It should be noted that the proposed approach still ignores grid identification and must remap each particle to each grid.

The second limitation is that the estimation of grids belonging to large and homogeneous grid regions usually has high uncertainty and low accuracy. For example, a particle that is predicted to be in the middle of a larger object can attach to any of its occupied grids. This can lead to ambiguous data correlations and ultimately higher uncertainty. To improve the estimation accuracy in uniform grid regions (i.e., in the middle of large objects with the same occupancy values), additional features from different sensors, larger patches, or more particles are needed.

\subsubsection{Memory-Efficient 3D Grid Mapping}\label{sec:memory-effi}
Memory is the major bottleneck for 3D occupancy mapping in large-scale scenes with small resolution. There are several explicit mapping representations, such as voxel, mesh, surfel, voxel hashing, Truncated Signed Distance Fields (TSDF) and Euclidean Signed Distance Fields (ESDF). The vanilla voxel occupancy grid map queries storage by index, which needs high memory loads, so it is not usual in mapping methods. Mesh stores surface information about obstacles Surfel consists of points and patches, which include radii and normal vectors. Voxel hashing is a memory-efficient improvement of vanilla voxel methods, which only divides voxel on the scene surface measured by the camera, and stores the voxel block on the scene surface in the form of a hash table to facilitate the query of voxel block. Octomap\cite{Octomap} introduces an efficient probabilistic 3D mapping framework based on octrees. Octomap iteratively divides the cube space into eight small cubes, with the large cube becoming the parent node and the small cube becoming the child node, and the octopus tree can continuously expand down until it reaches a minimum resolution, called a leaf node. Octomap uses probabilistic descriptions to easily update node status based on sensor data.

Continuous mapping algorithms is another alternative for computation- and memory-efficient 3D occupancy description with arbitrary resolution. Gaussian Process Occupancy Maps (GPOM) uses the modified Gaussian process as a non-parametric Bayesian learning technique that introduces dependencies between points on the map for continuity. Hilbert Maps\cite{HilbertMaps} projects the raw data into Hilbert space, where the logistic regression classifier is trained. BGKOctoMap-L\cite{BGKOctoMap} extends the traditional counting model CSM, and after smoothing it with a nuclear function, observations of surrounding voxels can be taken into account. AKIMap\cite{AKIMap} is based on BGKOctoMap, and the point of improvement is that the kernel function is no longer radial based, adaptively changing direction and adapting to boundaries. DSP-map\cite{DSP-MAP} generalizes particle-based map into continuous space and builds a continuous 3D local map adaptable for both indoor and outdoor applications. Broadly speaking, MLP structure in NeRF series is also an implicit continuous mapping for 3D geometry with almost no storage needs.

\subsection{Bird's-Eye View 2D Grid Representation}\label{sec:BEVRepre}
BEV grid is a common representation of obstacle detection for on-road vehicles. The basic technique for grid-centric perception is to map raw sensor information to BEV grid cells, which differ in mechanisms for different sensor modalities. LiDAR point clouds are naturally represented in 3D space, so there is a long-standing tradition of extracting point or voxel features on BEV maps\cite{PointRCNN,PV-RCNN}. Cameras are rich in semantic cues but lack geometric cues, which makes 3D reconstruction an ill-posed problem. The rising Vision BEV perception casts new lights on view transformation between perspective view to bird's eye view (PV2BEV).

\subsubsection{LiDAR-based Grid Mapping}\label{sec:lidar2bev}
Feature extraction of LiDAR point clouds follows the following paradigms: point, voxel, pillar, range view, or hybrid feature from above\cite{3DODSurvey}. This section focuses on the feature mapping of point clouds to BEV grids. 

\begin{figure}[ht!]
	\centering
	\includegraphics[width=0.3\textwidth]{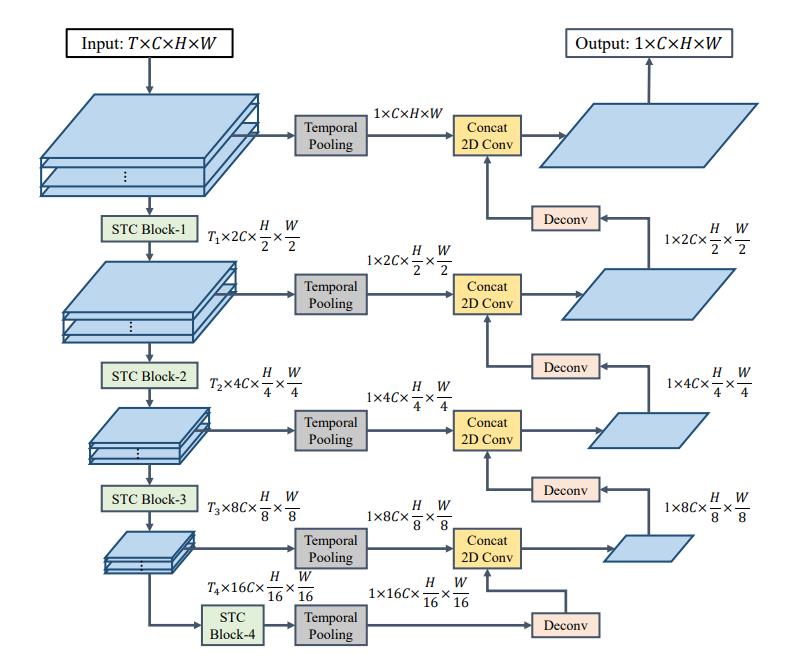}
	\caption{STPN\cite{MotionNet}: Spatial-temporal encoder which extracts multi-frame voxel features using only 2D CNN blocks, used in \cite{MotionNet,MotionSC,LiCaNet,LiCaNeXt}}
	\label{motionnet_stpn}
\end{figure}

LiDAR data collected in 3D space can be easily transformed to BEV and fused with information from multi-view cameras. The sparse and variable density of LiDAR point clouds renders CNNs inefficient. Some methods \cite{MV3D,AVOD,PIXOR} voxelize the point cloud into a uniform grid and encode each grid cell with hand-crafted features. MV3D\cite{MV3D}, AVOD\cite{AVOD} generates the BEV representation by encoding each grid with height, intensity, and density features. BEV representation in PIXOR\cite{PIXOR} is a combination of the 3D occupancy tensor and the 2D reflectance map, which keep the height information as channels. BEVDetNet\cite{BEVDetNet} further reduces BEV-based model latency to 2ms on Nvidia Xavier embedded platform. For advanced temporal tasks on grids, MotionNet proposes a novel spatial-temporal encoder STPN\cite{MotionNet} which aligns past point clouds to the current ego pose. The network design is shown in Fig. \ref{motionnet_stpn}. 

\begin{figure}[ht!]
	\centering
	\includegraphics[width=0.45\textwidth]{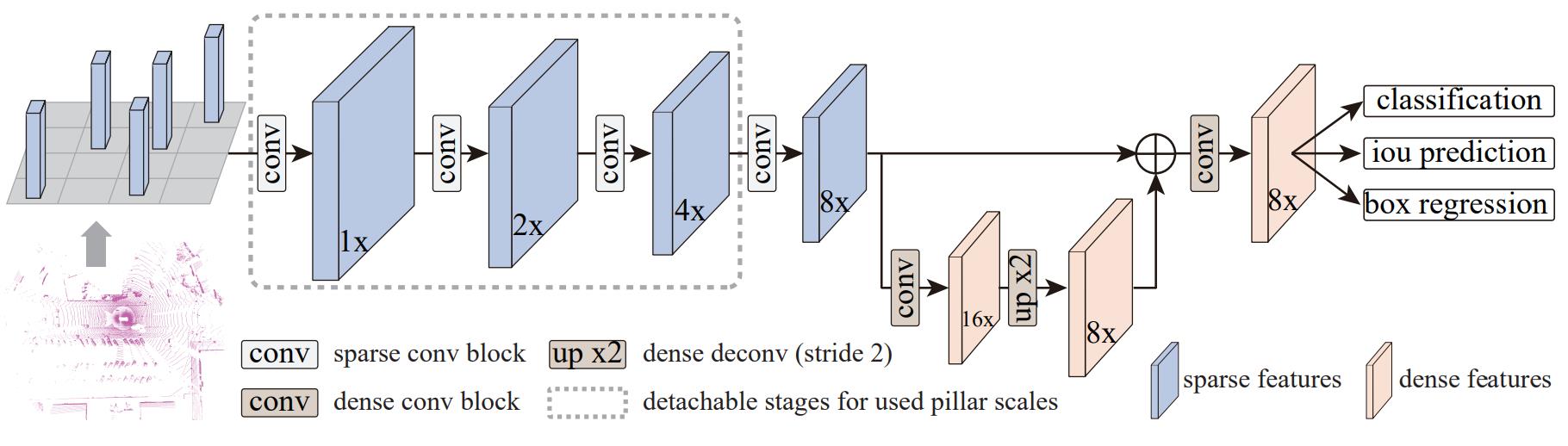}
	\caption{PillarNet\cite{PillarNet}: The pillar-based encoder proposed in PillarNet achieves state-of-the-art performance as well as super fast computation.}
	\label{pillarnet_pipeline}
\end{figure}

However, these fixed encoders are not successful in utilizing all the information contained within the point cloud. The learned features become a trend. VoxelNet\cite{VoxelNet} stacks voxel feature encoding (VFE) layers to encode point interactions within a voxel and generates a sparse 4D voxel-wise feature. And then VoxelNet uses a 3D convolutional middle layer to aggregate and reshape this feature and passes it through a 2D detection architecture. To avoid hardware-unfriendly 3D convolution, the pillar-based encoder in PointPillars\cite{PointPillars} and EfficientPillarNet\cite{efficientpillarnet} learns features on pillars of the point cloud. The features can be scattered back to the original pillar positions to generate a 2D pseudo-image. PillarNet\cite{PillarNet} further develops pillar representation by fusing densified pillar semantic features with spatial features in the neck module for final detection with orientation-decoupled IoU regression loss. The encoder for PillarNet\cite{PillarNet} is illustrated in Fig. \ref{pillarnet_pipeline}.

\subsubsection{Camera-based Grid Mapping}\label{sec:pv2bev}
As cameras provide information rich in semantic cues but lossy in geometric cues. Early researchers conduct 3D vision tasks on the perspective view (PV), which lacks enough geometry guidance. In recent years, vision BEV-based approaches achieve great success as BEV provides a structured coordinate encoding for pixel features from multi-view cameras. View projection from perspective view (PV) to bird's eye view (BEV) is a major issue for vision-centric grid perception.

\textbf{Geometry-based View Transformation}
Geometry-based methods utilize the natural geometric projection relationship between PV and BEV for transformation. These methods can be further divided into two categories: homograph-based and depth-based.  Homograph-based methods are efficient for learning road surface layouts for autonomous driving, due to their hard flat-ground assumption. Depth-based methods are more common in representing obstacles on the road and achieving state-of-the-art performance.

\textbf{Homograph-based methods.} Inverse Perspective Mapping (IPM)\cite{IPM} is the earliest and the most important work which uses a homography matrix to formulate the transformation under the constraint that the corresponding points lie on a horizontal plane. The homography matrix can be mathematically derived from intrinsic and extrinsic parameters of the camera. In addition to image projection, Cam2BEV\cite{Cam2BEV}, 3D-LaneNet\cite{3D-LaneNet} use IPM to transform feature maps for downstream perception tasks. Gu et al.\cite{HomoLoss} propose a Homography Loss for the training of the network. To reduce the distortion of the part above the ground plane caused by IPM, OGMs\cite{OGMs}, SHOT\cite{SHOT} use semantic information and MonoLayout\cite{MonoLayout} utilizes Generative Adversarial Network (GAN)\cite{GAN} to improve BEV features.

\begin{figure}[ht!]
	\centering
	\includegraphics[width=0.45\textwidth]{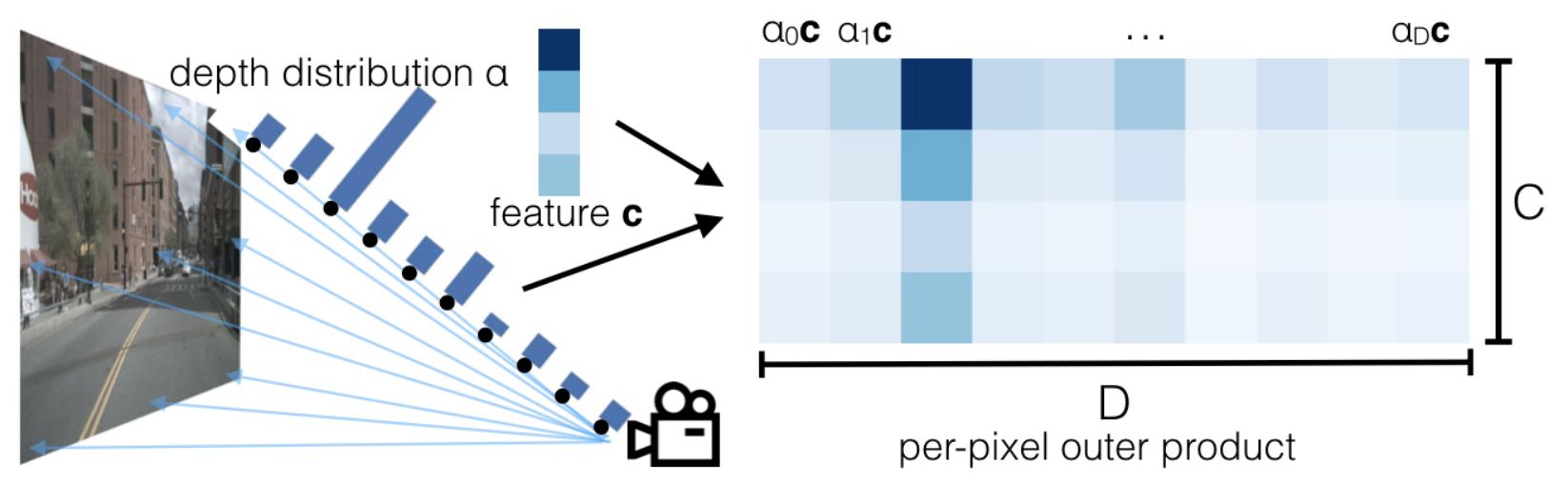}
	\caption{LSS\cite{LSS}: Probabilistic depth estimation in unprojecting images to BEV space, used in \cite{BEVDet,BEVDet4D,BEVerse,BEVFusion,BEVFusion2}}
	\label{lss_depthnet}
\end{figure}

\textbf{Depth-based methods.} These methods rely heavily on voxels, an explicit 3D representation compatible with the end-to-end design. As shown in Fig. \ref{lss_depthnet}, depth-based methods use the predicted depth distribution per grid to lift 2D features to 3D voxel space and then generate the BEV 2D grid representation. This lift process writes:
\begin{equation}
	\mathcal{F}_{3D}(x,y,z)=[\mathcal{F}^{*}_{2D}(u,v)\otimes \mathcal{D}^{*}(u,v)  ]_{x,y,z},
\end{equation}
where $(u,v)$ is pixel coordinate corresponding to $(x,y,z)$ in 3D space, $\mathcal{F}$ is feature, $\mathcal{D}$ is predicted depth distribution, * denotes one or more images, $\otimes$ denotes outer production. 

Depth is commonly formulated as a distribution. OFT\cite{OFT}, M2BEV\cite{M2BEV} assume uniform depth distribution, which means that all 3D features along a ray are the same. Some other depth-based methods\cite{CaDDN,DfM,LSS,BEVDepth,STS,MV-FCOS3D++} model the depth distribution with different learnable network structures and parameters.

Depth-based methods are usually supervised by labels from downstream tasks, but per-pixel depth supervision can also improve accuracy of depth estimation. CaDDN\cite{CaDDN}, DfM\cite{DfM}, BEVDepth\cite{BEVDepth} and STS\cite{STS} can optionally use depth supervision from LiDAR point clouds during the training stage. 

Some structures are able to capture stereo cues in ring-camera settings. MV-FCOS3D++\cite{MV-FCOS3D++} uses pre-trained 2D feature extractors for a few stereo cues. BEVDepth proposes a depth correction module to solve the spatial misalignment between the image feature map and the ground truth derived from LiDAR caused by the ego vehicle's motion. Combined with the monocular depth module in BEVDepth, BEVStereo\cite{STS} uses the geometrical correspondence between temporal dynamic stereo cameras to promote depth learning. 

Depth-based methods are still dominant in vision-centric perception. Recent works such as FIERY\cite{FIERY}, M2BEV\cite{M2BEV}, StretchBEV\cite{StretchBEV} and BEVerse\cite{BEVerse} are based on BEV representation derived from depth-based LSS\cite{LSS}.

\textbf{Fisheye BEV algorithm.} Compared to pinhole cameras, fisheye cameras have a broader field of view and higher distortion. Most fisheye detection algorithms are based on the perspective view. Wu et al.\cite{FPD_FPNet_TIV} propose a surround-view fisheye dataset FPD and BEV-perception multi-task model, FPNet. It has a shared feature extractor as well as 2D and 3D detection heads and per-pixel depth head.

% \begin{figure}[ht!]
% 	\centering
% 	\includegraphics[width=0.45\textwidth]{fig/bevdepth_depthnet.jpg}
% 	\caption{BEVDepth\cite{MotionNet}: Depth estimation module which considers context information in BEVDepth}
% 	\label{bevdepth_depthnet}
% \end{figure}

\textbf{Network-based View Transformation}

Network-based transformation methods utilize deep neural networks as a mapping function to model the view transformation and use the camera geometry implicitly. Recent methods can also be further divided into two categories: MLP-based and transformer-based. The former conducts the transformation in a forward way, while the latter constructs BEV queries and searches for corresponding features on PV maps through a cross-attention mechanism.

\textbf{MLP-based methods.} Multi-layer perceptron is employed to model the transformation from PV to BEV without camera parameters. Therefore, they avoid inductive biases from the calibrated camera system. VED\cite{VED} utilizes a variational encoder-decoder MLP network to predict BEV semantic occupancy grid map from a front-view monocular image in real-time. To use the information from different views to achieve surrounding sensing, VPN\cite{VPN} designs a view transformer module (VTM) with two-layer MLP to fuse each PV feature map into a BEV feature map. FishingNet\cite{FishingNet} transforms the camera features to BEV maps by modifying the VTM, and then fuse them with radar and LiDAR data for multi-modal perception and prediction. To predict more accurate BEV maps from monocular images, PON\cite{PON} and STA-ST\cite{STA-ST} include a feature pyramid network (FPN)\cite{FPN} to provide high-resolution features with rich context. The dense transformer layer in PON collapses PV features along height axis and expands them along the depth axis through MLP before resampling them to obtain BEV features in Cartesian coordinates. HDMapNet\cite{HDMapNet} puts forward a novel view transformer module that consists of MLP-based neural feature transformation and geometric projection. It can produce a vectorized local semantic map from images of the surrounding cameras. Since the projection of BEV features back to the PV map can form self-supervision, PYVA\cite{PYVA} includes a cross-view transformation module that utilizes an MLP-based cycle to retain the features related to view transformation. To reap the benefits of homograph-based methods and MLP-based methods, HFT\cite{HFT} consists of two branches to utilize geometric priors and global context respectively.

MLP-based methods tend to transform multi-view images individually and ignore the geometric potential brought by overlapping regions. These shortcomings make its application less widespread than subsequent transformer-based methods.

\textbf{Transformer-based methods.} Vaswani et al.\cite{Transformer} introduce transformer, an attention-based model which facilitates the development of natural language processing (NLP). Vision transformers have shown remarkable performance in various vision tasks due to its ability to capture long-range dependencies and global receptive fields. Compared to MLP-based methods, transformer-based methods view transformation have more impressive PV2BEV relation modeling ability. Query-based element extraction starts with 2D object detector DETR\cite{DETR}. 

To satisfy the requirements of grid-centric perception tasks, the granularity of the query in the transformer decoder should be finer and each query is position-encoded according to its spatial location in 3D space or BEV space. Tesla\cite{Tesla} first conducts the view transformation through cross-attention between dense BEV queries and multi-view image features. CVT\cite{CVT} points out that geometric reasoning of cross-attention transformers performs well and inferences faster than geometric models. So it employs cross-view transformers for view transformation, which use positional embeddings derived from calibrated camera intrinsics and extrinsics. In addition to an MLP-based cycled view projection mentioned above, PYVA's\cite{PYVA} transformation module also contains a cross-view transformer to correlate the features before and after view projection.

\begin{figure}[ht!]
	\centering
	\includegraphics[width=0.45\textwidth]{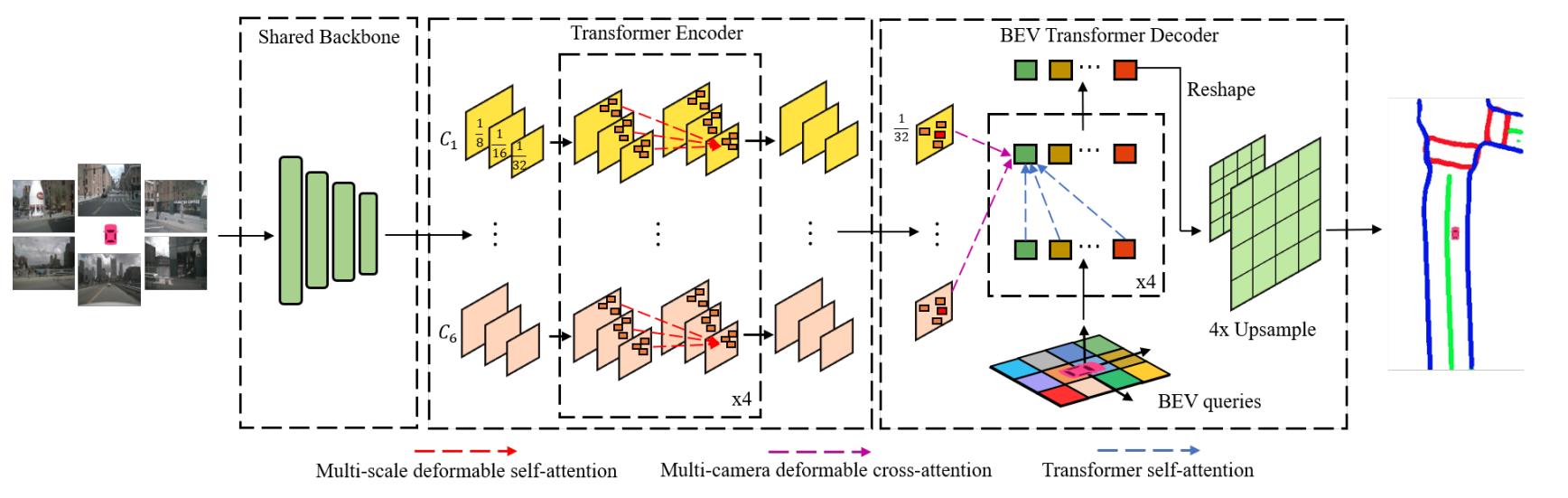}
	\caption{BEVSegFormer\cite{BEVSegFormer}: The transformer-based BEV segmentation model introduced in BEVSegFormer.}
	\label{bevsegformer_pipeline}
\end{figure}

To avoid high computation complexity in the transformer attention module, one approach is to use sparse attention. Deformable attention\cite{DeformableAttention} reduces the number of keys to avoid excessive attention computation by only focusing on relevant regions and capturing informative features. The view transformation module of many works use this sparse attention, such as BEVSegFormer\cite{BEVSegFormer} shown in Fig. \ref{bevsegformer_pipeline}, PersFormer\cite{PersFormer} and BEVFormer\cite{BEVFormer}. Note that the queries in BEVSegFormer directly predict the reference points on images, while PersFormer and BEVFormer rely on camera parameters to compute the reference points. Apart from deformable attention, several kinds of sparse attention are put forward. GKT\cite{GKT} unfolds kernel regions around the projected reference points and makes them interact with dense BEV queries to generate BEV representation. ViT-BEVSeg\cite{ViT-BEVSeg} uses simpler and plain vision transformer for patch embedding-based view transformation.

Another approach is to simplify the cross-attention computation with 3D geometry constraints. Image2Map\cite{Image2Map} uses camera geometry to dictate one-to-one correspondence between vertical scan lines in the image and polar rays in the BEV map to form an inter-plane attention mechanism. The view transformation is treated as a set of sequence-to-sequence translations. Similarly, PolarFormer\cite{PolarFormer} also employs this correspondence to simplify the global attention into column-wise attention and uses a polar alignment module to aggregate rays from multiple camera views to generate a structured Polar BEV feature map.

PETRv2\cite{PETRv2} is developed from PETR\cite{PETR} which contains sparse object queries for object-centric tasks. In order to form a unified framework for 3D perception, PETRv2 adds a set of BEV segmentation queries to support high-quality BEV segmentation, an important grid-centric task.

Geometric cues and temporal cues are both important to transformer-based methods. As geometric cues, calibrated camera parameters are utilized in many transformer-based methods to reduce memory/computation complexity or improve data efficiency. As mentioned above, \cite{CVT,PETRv2} generate 3D positional embedding derived from camera parameters to help the transformers learn view transformation. \cite{PersFormer,BEVFormer,GKT} rely on camera parameters to compute the reference points for feature sampling. \cite{Image2Map,PolarFormer} employs camera geometry to dictate the relationship between each vertical scanline in the image and the BEV features. However, with such geometric cues it is possible to introduce inductive biases contained in the calibrated camera system. Temporal cues which are used in \cite{BEVFormer,PolarFormer} can improve the performance of methods.

\textbf{Polar coordinate.} Common practices adopt Cartesian coordinate with evenly distributed rectangular grids. However, PolarBEV\cite{PolarBEV} argues that surrounding grids should be paid more attention than distant ones, so unevenly distributed polar rasterization can reduce computational budget. PolarDETR\cite{PolarDETR} models multi-view BEV detection transformer on Polar parameterizations. PolarFormer\cite{PolarFormer} extracts dense BEV features on polar coordinates and surpasses Cartesian coordinate counterparts in BEVFormer\cite{BEVFormer}. 

\subsubsection{Deep Fusion on Grids}\label{sec:deepfusion}

Multi-sensor multi-modality fusion has been a long-standing issue for automotive perception. Fusion frameworks are often categorized into early fusion, deep fusion, and late fusion. Among them, deep fusion has demonstrated the best performance in an end-to-end framework. Grid-centric representation serves as a unified feature embedding space for deep fusion among multiple sensors and agents.

\textbf{Multi-sensor Fusion}\label{sec:multisensorfusion}
Different sensors are inherently complementary: Cameras are geometry-loss but semantics rich, while LiDARs are semantics-loss but geometry rich. Radars are geometry and semantics sparse but robust to different weather conditions. Deep fusion fuses latent features across modalities and compensates for the limitations of each sensor. 

\begin{figure}[ht!]
	\centering
	\includegraphics[width=0.45\textwidth]{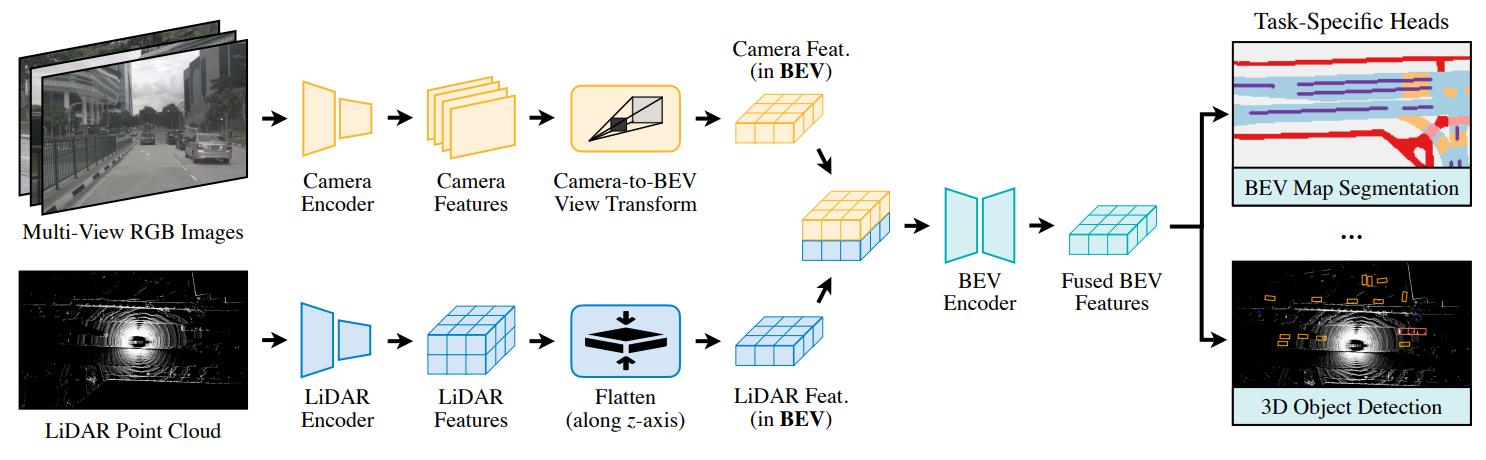}
	\caption{BEVFusion\cite{BEVFusion}: A simple and unified BEV feature-level fusion framework for LiDARs, cameras and radar points.}
	\label{bevfusion_pipeline}
\end{figure}

\textbf{LiDAR-camera fusion.} Some methods perform the fusion operation at a higher 3D level and support feature interaction in 3D space. UVTR\cite{UVTR} samples features from the image according to predicted depth scores and associates features of point clouds to voxels according to the accurate position. Thus, the voxel encoder for cross-modality interaction in voxel space can be introduced.  AutoAlign\cite{AutoAlign} designs a cross-attention feature alignment module (CAFA) to enable the voxelized feature of point clouds to perceive the whole image and perform feature aggregation. Instead of learning the alignment through the network in AutoAlign\cite{AutoAlign}, AutoAlignV2\cite{AutoAlignV2} includes a cross-domain DeformCAFA and employs the camera projection matrix to obtain the reference points in the image feature map. FUTR3D\cite{FUTR3D} and TransFusion\cite{Transfusion} fuses features based on attention mechanism and queries. FUTR3D employs a query-based modality-agnostic feature sampler(MAFS) to extract multi-modal features according to 3D reference points. TransFusion relies on LiDAR BEV features and image guidance to generate object queries and fuses these queries with image features. A simple and robust approach is to unify fusion on BEV features. Two implementations of BEVFusion\cite{BEVFusion,BEVFusion2}, shown in Fig. \ref{bevfusion_pipeline} unify features from multi-modal inputs in a shared BEV space. DeepInteration\cite{DeepInteration} and MSMDFusion\cite{MSMDFusion} designs multi-model interaction in BEV space and voxel space to better align spatial features from different sensors.

\textbf{Camera-radar fusion.} Radar sensors are originally designed for Advanced Driving Assistance System (ADAS) tasks, so their accuracy and density are insufficient for use in high-level autonomous driving tasks. OccupancyNet\cite{OccupancyNet} and NVRadarNet\cite{NVRadarNet} only use radars to perform real-time obstacle and free space detection. Camera-radar fusion is a promising low-cost perception solution that supplements semantics to radar geometry. Simple-BEV\cite{Simple-BEV}, RCBEVDet\cite{RCBEVDet}, and CramNet\cite{CramNet} have investigated different approaches for radar feature expression on BEV and fusion with vision BEV features. RCBEVDet\cite{RCBEVDet} process the multi-frame aggregated radar point clouds with a PointNet++\cite{PointNet++} network. CramNet\cite{CramNet} sets the camera features as query and radar features as value to retrieve radar features along the pixel ray in 3D space. Simple-BEV\cite{Simple-BEV} voxelizes multi-frame radar point clouds as a binary occupancy image and uses meta-data as additional channels. RRF\cite{RRF} yields a 3D feature volume from each camera by projection and sampling and then concatenates a rasterized radar BEV map. It finally gets a BEV feature map by reducing the vertical dimension.

\textbf{Lidar-camera-radar fusion.} LiDAR, radar, and camera fusion is a robust fusion strategy for all weathers. RaLiBEV\cite{RaLiBEV} adopts an interactive transformer-based bev fusion that fuses LiDAR point clouds and radar range azimuth heatmaps. FishingNet\cite{FishingNet} uses top-down semantic grids as a common output interface to conduct late fusion of the LiDAR, radars and cameras and performs short-term prediction of semantic grids. 

\textbf{Multi-agent Fusion}\label{sec:multiagentfusion}
Recent works on grid-centric perception are mostly based on single-agent systems, which have limitations in complex traffic scenes. Advancements in Vehicle-to-Vehicle(V2V) communication technologies enable vehicles to share their sensory information. CoBEVT\cite{CoBEVT} is the first multi-agent multi-camera perception framework that can generate BEV segmented maps cooperatively. In this framework, the ego vehicle geometrically warps the received BEV features according to the pose of the sender and then fuses them using a transformer with fused axial attention (FAX). Dynamic occupancy grid map (DOGM) also shows the capacity of reducing uncertainty in the fusion platform for multi-vehicle cooperative perception\cite{CoopDOGM,MVMS-OGM, GevBEV}.

\subsection{Related Perception Tasks}\label{sec:relatedtasks}

\subsubsection{Simultaneous Localization and Mapping}\label{sec:slam}
The technique of Simultaneous Localization and Mapping (SLAM) is essential for mobile robots to navigate in an unknown environment. SLAM is highly related to geometry modeling. In the LiDAR SLAM field, High Order CRFs\cite{HighOrderCRF} proposes an incrementally constructed 3D scrolling OGM for efficiently representing large-scale scenarios. SUMA++\cite{SUMA++} directly employs RangeNet++\cite{RangeNet++} for LiDAR segmentation, semantic ICP\cite{ICP} only for stationary environments, and semantic-based dynamic filter for surfel map reconstruction.  In the visual SLAM field, ORB-SLAM\cite{ORB-SLAM} stores maps with points, lines or planes. Diving the space into discrete grids is commonly used in dense and semantic mapping algorithms\cite{VolumetricCRF}. An new trend is to combine neural fields with SLAM with two advantages: NeRF models manipulate raw pixel value directly without feature extraction; NeRF model can differentiate both implicit and explicit representation, leading to a full-dense optimization of 3D geometry. NICE-SLAM\cite{NICE-SLAM} and NeRF-SLAM\cite{NeRF-SLAM} are able to generate dense, hole-free maps. NeRF-SLAM generates a volumetric NeRF whose dense depth loss is weighted by the depth's marginal covariance.

\subsubsection{Map Element Detection}\label{sec:mapdet}
\begin{figure}[ht!]
	\centering
	\includegraphics[width=0.45\textwidth]{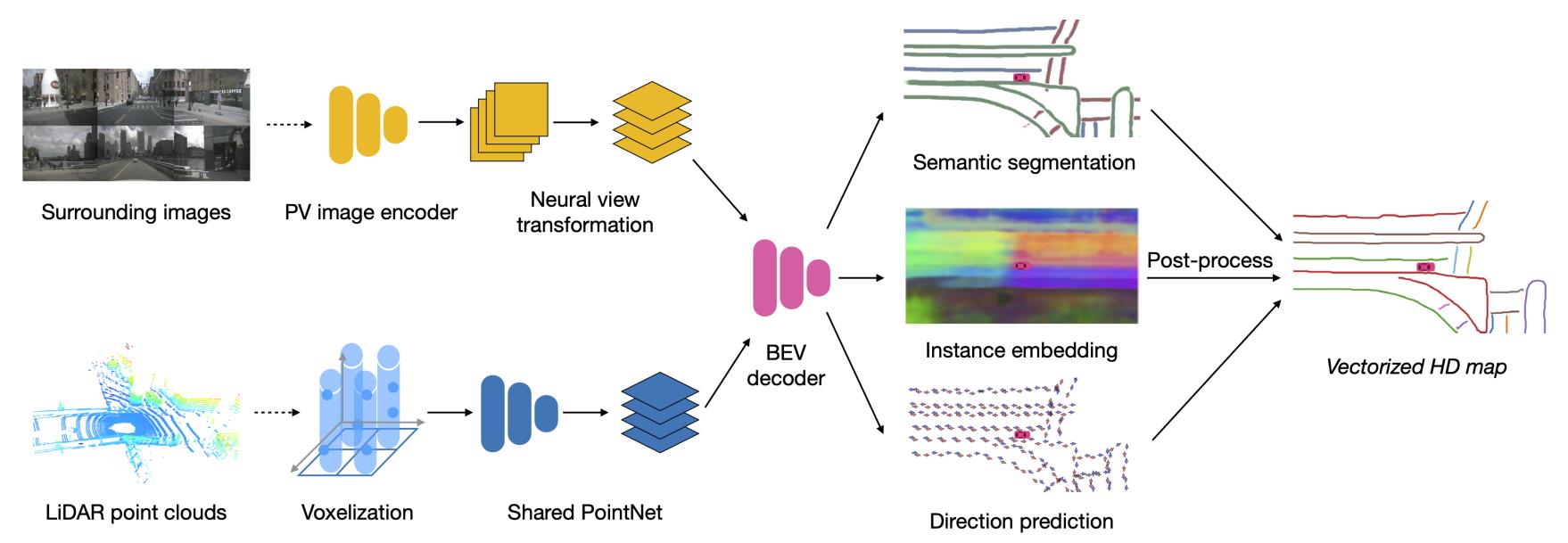}
	\caption{HDMapNet\cite{HDMapNet}: End-to-end map element and direction extraction network, where the final vectorization is based on post-processing techniques.}
	\label{hdmapnet_pipeline}
\end{figure}

Detecting map elements is a crucial step in the production of high-definition maps. Traditional global map construction requires an offline global SLAM with a globally consistent pointcloud and centermeter-level localization. In recent years, a novel approach has been an end-to-end online learning approach based on BEV segmentation and post-processing techniques for local map learning, and then connecting local maps in different frames to produce a global high definition map\cite{HDMapNet}. The entire pipeline is depicted in Fig. \ref{hdmapnet_pipeline}.

\begin{figure}[ht!]
	\centering
	\includegraphics[width=0.45\textwidth]{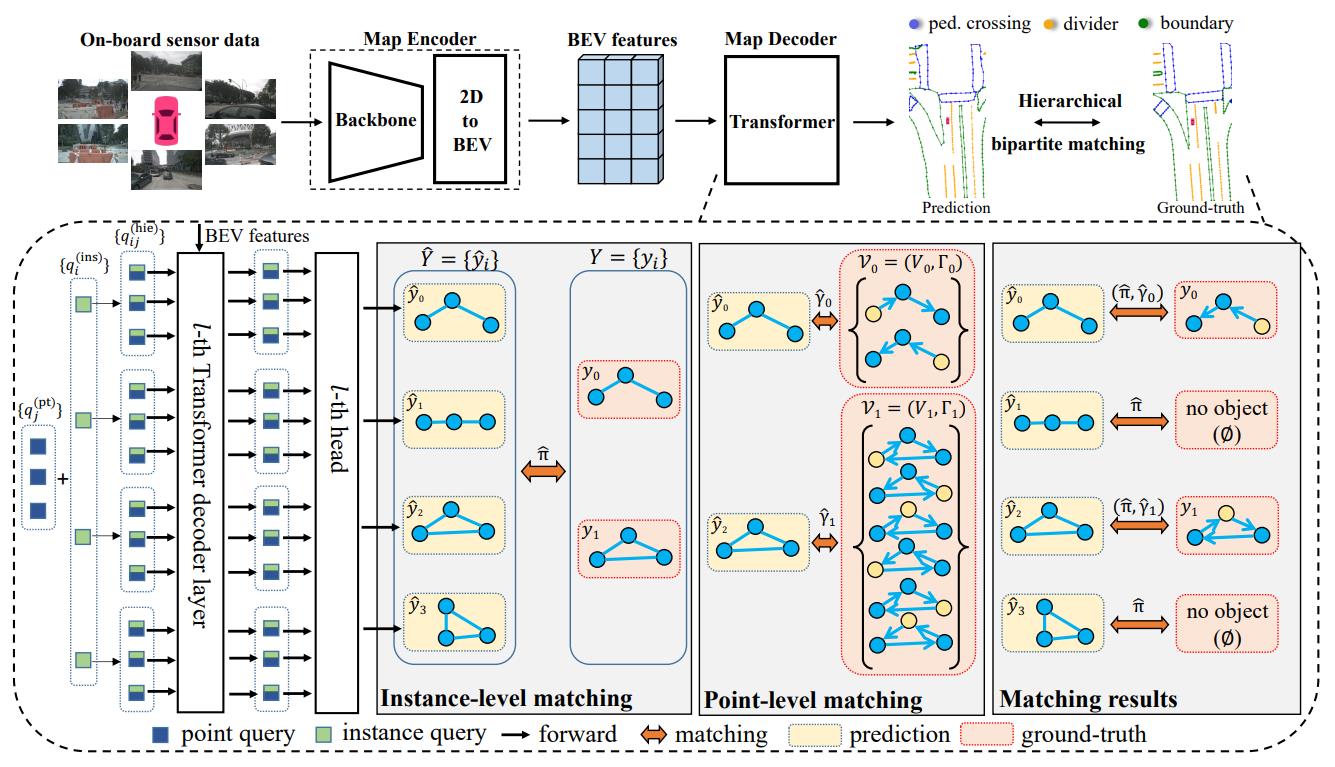}
	\caption{MAPTR\cite{MAPTR}: End-to-end vector map extraction which represents vectors as queries and finds correlation in a transformer-based framework.}
	\label{maptr_pipeline}
\end{figure}

Typically, vectorized map elements are required for HD map-based applications, such as localization or planning. In HDMapNet\cite{HDMapNet}, vectorized map elements can be generated through post-processing of BEV segmentation of map elements; however, end-to-end approaches have recently gained favor. End-to-end pipelines consist of feature extraction of on-board LiDARs and cameras introduced in Section \ref{sec:BEVRepre} and transformer-based heads which regress vector element candidates as queries and interact with values in the BEV feature map. STSU\cite{STSU} extracts road topology from a structured traffic scene by utilizing a Polyline-RNN\cite{Polyline-RNN} that extracts initial point estimates to form the centerline curves. VectorMapNet\cite{VectorMapNet} directly predicts a sparse set of polylines primitives to represent the geometry of HD maps. InstaGram\cite{InstaGram} proposes a hybrid architecture with CNN and graph neural network (GNN), which extracts vertex locations and implicit edge maps from BEV features. A GNN is utilized to vectorize and connect the HD map's elements. As depicted in Fig. \ref{maptr_pipeline}, MAPTR\cite{MAPTR} proposes a hierarchical query embedding scheme to encode instance-level and point-level bipartite matching for map element learning.

\subsubsection{Multi-task Models}\label{sec:multitasking}
Many researches reveal that predicting geometric, semantic and temporal tasks together in a multi-task model improves accuracy of each respective model. Recent advances handles more perception tasks other than grid-centric tasks in one base framework. A unified framework on BEV grids is efficient for an automotive perception system, this section introduces some commonly used multi-task learning frameworks. 

% Fig. \ref{chrono_multitask_model} provides a chronological review for existing multitask models and their respective tasks. 

\textbf{Joint BEV Segmentation and Prediction}\label{sec:multitasking-segmotion}
Accurate recognition of moving objects in BEV grids is an important prerequisite for BEV motion prediction. Therefore, past practices have demonstrated that accurate semantic recognition helps motion and velocity estimation. Common practices includes a spatial-temporal feature extraction backbone and task-specified heads, segmentation head to classify to which class the grid belongs, state head to classify stationary or dynamic grid, instance head which can predict offset of each grid to an instance center and motion head to predict short-term motion displacement. Vision-centric BEV models usually jointly optimize instances' category, location and coverage, FIERY\cite{FIERY} introduces uncertainty loss\cite{UncertintyLoss} to balance the weight of segmentation, centerness, offset and flow losses.

% \begin{figure}[ht!]
% 	\centering
% 	\includegraphics[width=0.45\textwidth]{fig/motionnet_pipeline.jpg}
% 	\caption{MotionNet\cite{MotionNet}: The first spatial-temporal networks on multi-task for BEV segmentation and motion.}
% 	\label{motionnet_pipeline}
% \end{figure}
% \begin{figure}[ht!]
% 	\centering
% 	\includegraphics[width=0.45\textwidth]{fig/be-sti_pipeline.jpg}
% 	\caption{BE-STI\cite{BESTI}: BE-STI mainly improves the multi-task model by introducing TeSE and SeTE modules into voxel encoder.}
% 	\label{be-sti_pipeline}
% \end{figure}

\textbf{Comparison with LiDAR and camera-based BEV Segmentation and Motion.} An apparent difference is that LiDAR models estimate only grids accessible to laser scans. In other words, LiDAR-based methods have no completion ability for unobserved grid areas, or unobserved parts of dynamic objects. On the contrary, camera-based methods has techniques like probabilistic depth in LSS\cite{LSS} which can infer some kinds of occluded geometry behind observations. Hallucinating Beyond Observation\cite{Hallucinating-Beyond-Observation} is an example practice for inferring 3D shape of vehicles from images. Another difference is the generalization towards open-world unknown obstacles. MotionNet\cite{MotionNet} states that although trained on closed-set labels, MotionNet has the ability to predict motion of unknown labels which are all categorized into the 'other' class. However, camera-based methods are strict to classify well-defined semantics such as vehicle and pedestrian. Adaptability to open-world semantics of cameras remains an open question.

\begin{figure}[ht!]
	\centering
	\includegraphics[width=0.45\textwidth]{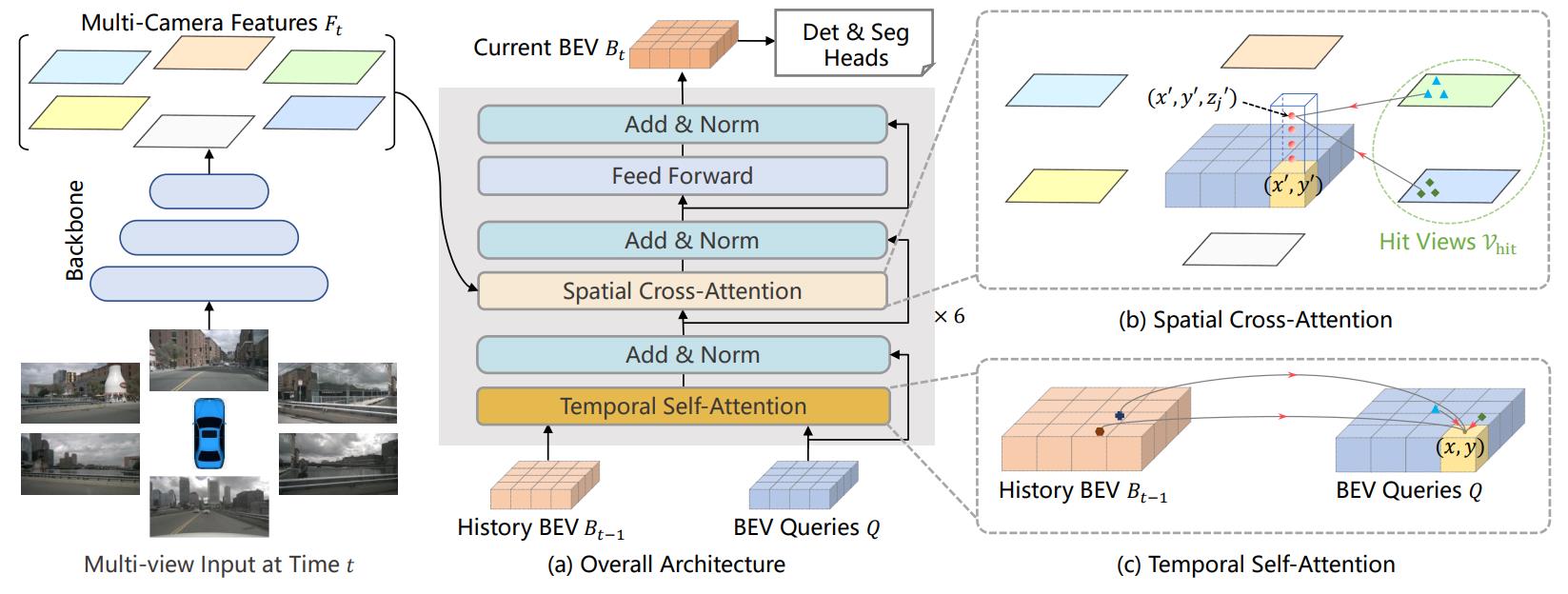}
	\caption{BEVFormer\cite{UniFormer}: Typical multi-task tasks based on BEV features. BEVFormer uses deformable-detr head for detection and panoptic segformer head for lane segmentation.}
	\label{bevformer_pipeline}
\end{figure}

\textbf{Joint 3D Object Detection and BEV Segmentation}\label{sec:multitasking-detseg}
Joint 3D object detection and BEV segmentation is a popular combination which handles perception of dynamic objects and static road layout in one unified framework. It is also one of the tracks held by SSLAD2022 workshop challenge\footnote{\url{https://sslad2022.github.io/pages/challenge.html}}. Given a shared BEV feature representation, common prediction heads for object detection are center head introduced in CenterPoint\cite{CenterPoint} and DETR head introduced in Deformable DETR\cite{DeformableDETR}, and common heads for segmentation are simple lightweight convolutional head (e.g.) and SegFormer\cite{SegFormer} or Panoptic SegFormer\cite{PanFormer} in BEVFormer\cite{BEVFormer}, or can easily extend to more complicated segmentation techniques. The pipeline of BEVFormer is shown in Fig. \ref{bevformer_pipeline}. MEGVII proposes the first place solution\cite{megvii_tech_report_sslad_multitasking} in SSLAD2022 multi-tasking challenge. They propose a multi-modal multi-task BEV model as a base. The model is pretrained on ONCE dataset and finetuned on AutoScenes dataset with techniques such as semi-supervised label correction and module-wise expotional moving average (EMA). 

\textbf{Multi-tasking for More Tasks}\label{sec:multitasking-more}
Recent researches places more major perception tasks in a BEV-based multi-tasking framework. BEVerse\cite{BEVerse} shows a metaverse of BEV features with 3D object detection, road layout segmentation and occupancy flow prediction. Perceive Interact Predict\cite{PerceiveInteractPredict} conducts end-to-end trajectory prediction based on interaction with map elements extracted online with shared BEV features. UniAD\cite{UniAD} is a comprehensive integration of object detection, tracking, trajectory prediction, map segmentation, occupancy and flow prediction, and planning, all in one vision-centric end-to-end framework. For more stable performance, UniAD is trained in two stages, tracking and mapping in the first stage and the whole model in the second stage.

% TODO: Add joint learning policy  Pareto is a promising MTL strategy

\textbf{Analysis: Antagonism in multitask BEV models.} 
A unified BEV feature representation and task-specified prediction heads compose an efficient framework design which is popular for industrial application. There remains a concern, whether the shared backbone strengthens each respective task. Joint BEV Segmentation and Motion researches\cite{BESTI} report a positive influence of multi-tasking: better segmentation leads to better motion prediction. However, most joint BEV detection and segmentation models\cite{BEVFormer,PerceiveInteractPredict,megvii_tech_report_sslad_multitasking} report the antagonism between two tasks. A reasonable explanations is that these two tasks are not relevant as they require features in different height, on the ground surface and above the ground. How shared BEV feature can generalizes well to fit each task needs adaption of specified feature maps remains an under-explored question.

\section*{Acknowledgments}
This work was supported in part by the National Natural Science Foundation of China under Grants (52394264, 52372414, U22A20104). This work was sponsored by Tsinghua-Zongmu Technology Joint Research Center.

\bibliographystyle{IEEEtran}
\bibliography{IEEEabrv, ref/ref}

\end{document}